%% file: main.tex
\documentclass[10pt,twocolumn,letterpaper]{article}

\usepackage[pagenumbers]{cvpr}     
\input{preamble}
\definecolor{titleblue}{HTML}{004182}
\usepackage[pagebackref,breaklinks,colorlinks,allcolors=titleblue]{hyperref}
\usepackage{algorithm}
\usepackage{amssymb}
\usepackage{amsmath}
\usepackage{algpseudocode}
\usepackage{booktabs}
\usepackage{color}
\usepackage[table,xcdraw]{xcolor}
\usepackage{multirow}
\usepackage{arydshln}
\usepackage{graphicx}
\usepackage{capt-of}
\usepackage{fontawesome5}
\algnewcommand{\Output}{\item[\textbf{Output:}]}
\usepackage{caption}
\usepackage[most]{tcolorbox}
\usepackage{tikz}

\newtcolorbox{paperabstract}{
  enhanced,
  colback=white,
  colframe=titleblue,
  boxrule=0.5pt,
  arc=3mm,
  left=5mm,
  right=5mm,
  top=4mm,
  bottom=4mm,
  before skip=0pt,
  after skip=0pt
}

\author{
Jiahua Xu$^{1}$ \quad
Yijie Fang$^{1}$ \quad
Nannan Wang$^{1,*}$ \quad
Xunhao Lin$^{1}$ \quad
Li Cao$^{2}$ \quad
Lei Hao$^{3}$ \\ 
Mingrui Zhu$^{1}$ \quad
Fei Gao$^{1}$ \quad
Xinbo Gao$^{1}$ \quad\\
$^{1}$State Key Laboratory of Integrated Services Networks,
Xidian University, Xi'an, China \\
$^{2}$Technology Planning Dept of the Computing Technology Research Institute, Huawei, ShangHai, China \\
$^{3}$Huawei Noah's Ark Institute, Huawei, Shenzhen, China 
}

\begin{document}

\onecolumn
\thispagestyle{plain}

\noindent
{\color{titleblue}\rule{\textwidth}{1.2pt}}
\par

{\centering
  \rmfamily\LARGE\bfseries
  Any Forcing: Any-Step Generality and One-Step Stability \\ for Autoregressive Video Generation\par
}

\vspace{0.08cm}

{\color{titleblue}
\noindent\rule{\textwidth}{0.6pt}\par}

\vspace{0.25cm}

\begin{center}
  {\rmfamily\normalsize\bfseries
  Jiahua Xu$^{1}$ \quad
  Yijie Fang$^{1}$ \quad
  Nannan Wang\textsuperscript{1,\textcolor{titleblue}{\faEnvelopeOpenText}} \quad
  Xunhao Lin$^{1}$ \quad
  Li Cao$^{2}$ \quad
  Lei Hao$^{3}$\\[3pt]
  Mingrui Zhu$^{1}$ \quad
  Fei Gao$^{1}$ \quad
  Xinbo Gao$^{1}$
  \par}

  \vspace{0.15cm}

  {\rmfamily\small
  $^{1}$State Key Laboratory of Integrated Services Networks,
  Xidian University, Xi'an, China\\[2pt]
  $^{2}$Technology Planning Dept of the Computing Technology
  Research Institute, Huawei, Shanghai, China\\[2pt]
  $^{3}$Huawei Noah's Ark Institute, Huawei, Shenzhen, China
  \par}
\end{center}

\vspace{0.3cm}

\begin{paperabstract}
    {\centering
      \rmfamily\large\bfseries\color{titleblue}
      Abstract\par
    }
    \vspace{0.2cm}

  {\small
    \input{sec/0_abstract}
    \par
  }

  \vspace{0.3cm}

  {\footnotesize\centering
  \textcolor{titleblue}{\faGlobe}\
  \textbf{Project Page:}
  \href{https://medcx.github.io/Any-Forcing/}{Any Forcing}
  \quad\quad
  \textcolor{titleblue}{\faEnvelopeOpenText}\ \textbf{Correspondence:}
  Nannan Wang
  (\href{mailto:nnwang@xidian.edu.cn}{E-mail})
  \par}
\end{paperabstract}

\vspace{0.4cm}

\noindent
\begin{minipage}{\textwidth}
  \centering
  \includegraphics[width=\linewidth]{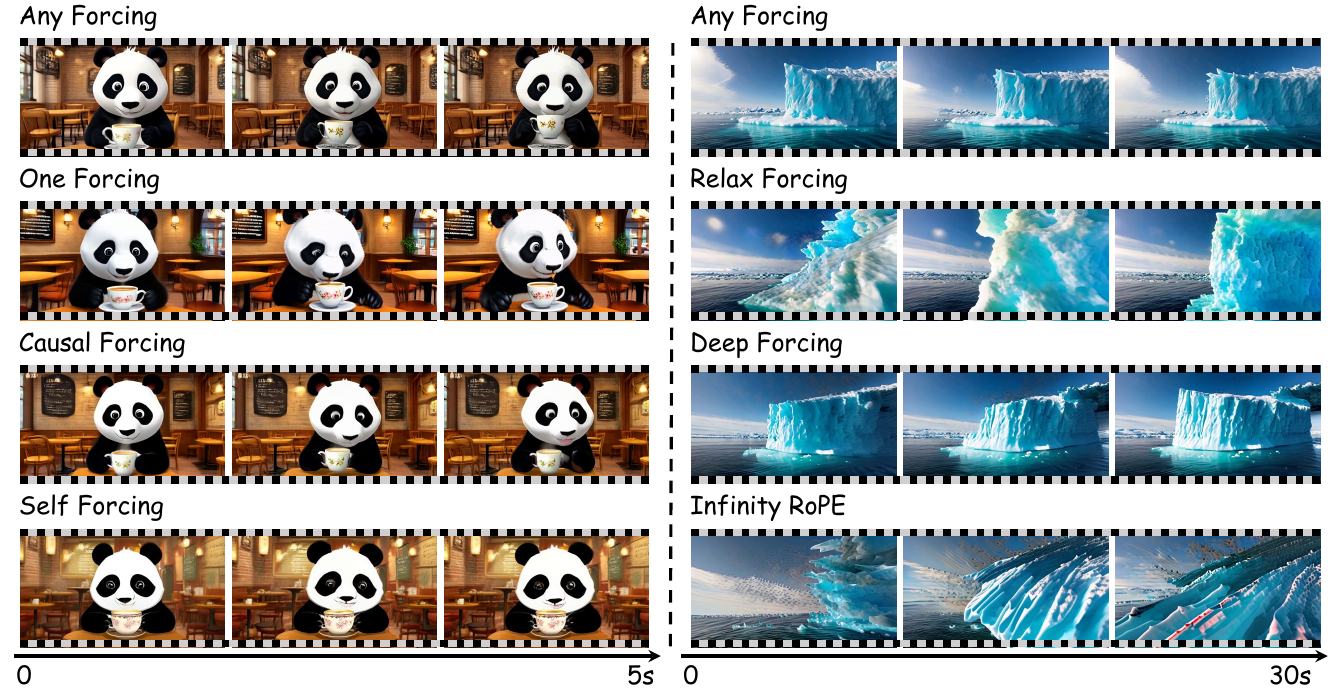}

  \captionsetup{
    font=small,
    labelfont=bf,
    labelsep=space,
    justification=justified,
    singlelinecheck=false,
    skip=6pt
  }

  \captionof{figure}{
    \textbf{Visual comparison of short- and long-video generation.}
    Existing methods exhibit degraded visual quality and suffer
    from long-horizon error accumulation under one-step sampling.
    Any Forcing achieves stable one-step generation and
    long-horizon consistency.
  }
  \label{fig:video_teaser}
\end{minipage}

\twocolumn
\input{sec/1_intro}

\input{sec/2_work}
\input{sec/3_method}
\input{sec/4_details}

\input{sec/5_results}

\input{sec/6_conclusion}

{
    \small
    \bibliographystyle{ieeenat_fullname}
    \bibliography{main}
}


\input{sec/supplementary}
\end{document}

%% file: sec/0_abstract.tex
Autoregressive video generation enables efficient inference through streaming prediction. However, existing methods are typically restricted to fixed sampling steps, limiting the flexibility to balance generation quality and efficiency, while one-step generation suffers from degraded visual quality and error accumulation over long sequences, leading to content drift and generation collapse. To address these issues, we propose \textbf{Any Forcing}, a unified framework for achieving \textbf{Any-Step Generality} and \textbf{One-Step Stability}. Specifically, we introduce a unified adversarial discriminator that jointly performs real-distribution alignment, inter-step consistency alignment, and first-frame quality enhancement, enabling flexible sampling while preserving one-step generation quality. Furthermore, we propose a training-free adaptive long-term memory selection mechanism that selects historical states based on global semantic relevance and local value contribution, while remapping their temporal positions to preserve temporal relationships within the memory cache, thereby enabling stable one-step long-video generation. Experiments demonstrate that Any Forcing achieves high-quality and stable one-step autoregressive video generation, achieving 20.9 FPS and 24.7 FPS for short and long video generation on a single H100 GPU, respectively.

%% file: sec/1_intro.tex
\section{Introduction}
\label{sec:introduction}

Autoregressive video generation~\cite{yin2025causvid,huang2026self,zhu2026causal} provides a paradigm for streaming prediction by progressively generating future content while leveraging previously generated frames to model temporal dependencies. Compared with bidirectional DiT-based video generation models~\cite{wan2025,kong2024hunyuanvideo,zheng2024open}, autoregressive generation can reuse historical attention features to reduce redundant computation and continuously incorporate newly arriving control signals during inference, making it particularly suitable for real-time video generation and long-horizon video extrapolation.

To further improve the efficiency of autoregressive video generation, recent studies have employed distillation and self-rollout training to compress multi-step sampling into few-step or even one-step generation~\cite{huang2026self,zhu2026causal,feng2026oneforcing,liu2026rolling,zhang2026codmd}.
However, these methods are still designed for fixed sampling steps, limiting their generality under different inference budgets. Although methods such as ASD~\cite{yang2026towards} improve compatibility across sampling steps by enforcing inter-step consistency, relying solely on student self-supervision provides limited guidance under extremely few-step settings, particularly for one-step generation, often resulting in degraded generation quality. Therefore, achieving both generality across sampling steps and high-quality one-step generation remains a key challenge for efficient video generation.

\begin{figure}[t]
    \centering
    \includegraphics[width=0.48\textwidth]{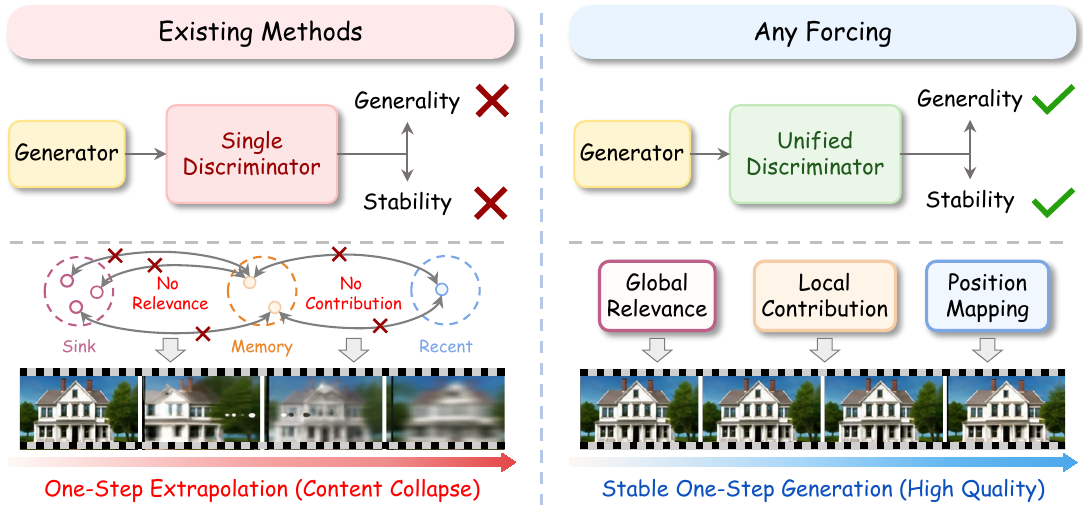}
    \caption{Motivation of Any Forcing. Existing methods struggle to balance sampling-step generality and one-step quality, while insufficient global relevance or local contribution undermines one-step stability in long-video generation. Any Forcing addresses these challenges through a unified adversarial discriminator and adaptive long-term memory selection mechanism.}
    \label{fig:motivation}
\end{figure}

Moreover, existing self-rollout-based methods are typically trained within limited temporal horizons, leaving long-term error accumulation insufficiently addressed. Recent long-video extrapolation methods~\cite{yesiltepe2026infinity,yi2025deep,liu2026rolling,zhao2026relax} extend temporal context through positional extrapolation and KV-cache management. However, as shown in Fig.~\ref{fig:motivation}, these methods struggle to effectively establish relationships among key components, resulting in rapid error accumulation under extremely few-step sampling, particularly in the one-step regime, and ultimately leading to content drift and visual collapse. Therefore, explicitly establishing relationships among key historical components is essential for stable one-step long-video generation.

To address the above challenges, we propose Any Forcing, which aims to achieve both \emph{Any-Step Generality} and \emph{One-Step Stability}. Specifically, during training, we introduce a unified adversarial discriminator that jointly enforces three complementary objectives: real-distribution alignment to promote distributional consistency between generated and real videos, thereby enhancing visual quality, inter-step consistency alignment to align predictions across different sampling steps, and first-frame enhancement to provide a stronger initialization for subsequent autoregressive generation. This unified adversarial design enables a single model to accommodate different sampling configurations while maintaining strong generation quality under one-step generation. Meanwhile, we introduce a training-free \emph{adaptive long-term memory selection} mechanism for stable one-step long video generation during inference, which maintains a fixed global sink for content consistency, recent local context for temporal continuity, and a dynamic long-term memory pool for long-term dependencies. We explicitly establish relationships between historical frames and both the global sink and recent context, jointly evaluating their global semantic relevance and local value contribution to guide long-term memory selection. Temporal position mapping is further applied to preserve temporal relations within the memory cache. Together, these strategies enable flexible sampling and stable one-step autoregressive video generation, achieving \emph{20.9 FPS and 24.7 FPS} for short and long videos on a single H100 GPU, respectively.

The main contributions are summarized as follows:

\begin{itemize}
    \item We propose a novel framework, Any Forcing, to achieve both Any-Step Generality and One-Step Stability for autoregressive video generation.
    
    \item A unified adversarial discriminator is introduced to achieve diverse sampling steps while maintaining strong one-step generation quality.
    
    \item We propose a training-free long-term memory selection mechanism to mitigate error accumulation in one-step long-video generation.
    
    \item Extensive experiments on short and long video generation demonstrate that Any Forcing achieves high-quality and stable one-step autoregressive video generation.
\end{itemize}

%% file: sec/2_work.tex
\section{Related Work}

\subsection{Diffusion Distillation}

Diffusion distillation~\cite{he2026one,he2026mixture,wang2024sinsr,cheng2026twinflow,salimans2022progressive} compresses multi-step diffusion models into few-step generators through three main paradigms. Consistency distillation~\cite{song2023consistency,lu2025simplifying,zheng2026large} enforces prediction consistency along teacher trajectories, enabling efficient sampling. Distribution Matching Distillation (DMD)~\cite{yin2024one,yin2024improved,nie2026transition,yuan2026helios,you2026adaptive} aligns real and generated distributions through score differences, with recent extensions to autoregressive video generation~\cite{huang2026self,zhu2026causal,yin2025causvid}. Adversarial distillation~\cite{cheng2026phased}, including ADD~\cite{sauer2024adversarial}, APT~\cite{lin2025diffusion} and AAPT~\cite{lin2026autoregressive}, leverages discriminators to improve generation fidelity through adversarial training. However, existing methods struggle to jointly support flexible sampling and stable one-step generation.

\subsection{Autoregressive Video Generation}

Autoregressive video generation~\cite{deng2025autoregressive,lin2026autoregressive,chen2026context,zou2026hiar,zhang2026reward,xue2026ring,zhuang2026self} enables efficient streaming prediction conditioned on historical context. Early studies~\cite{yan2021videogpt,kondratyuk2023videopoet} rely on discrete tokens, while recent approaches~\cite{yin2025causvid,huang2026self,zhu2026causal} adopt causal diffusion and distillation for efficient generation. Subsequent studies~\cite{yang2026towards,feng2026oneforcing,ma2026flex,lu2026reward} explore few-step and flexible sampling. For long-video generation, LongLive, Infinity RoPE, and Deep Forcing~\cite{yang2026longlive,yesiltepe2026infinity,yi2025deep} extend temporal context through memory management and positional remapping. However, maintaining stable one-step generation over long sequences remains challenging. Because generated frames condition subsequent predictions, errors can accumulate over time, leading to content drift and structural degradation. This motivates memory strategies that preserve both long-range semantic information and recent temporal dependencies within a limited context budget.

%% file: sec/3_method.tex
\section{Preliminary}

\noindent\textbf{Notation.}
We use bold italic lowercase letters
$\boldsymbol{x}_0^{\mathrm{real}}$ and
$\widehat{\boldsymbol{x}}_0^{(k)}$ to denote real video
latents and student-generated video latents obtained
with $k$ denoising steps, respectively.
The subscript $0$ indicates clean latents, and
$\boldsymbol{x}_0^{\mathrm{real}}
=(\boldsymbol{x}_0^1,\ldots,\boldsymbol{x}_0^F)$
contains $F$ latent frames.
Moreover, $t$ denotes the diffusion timestep, while
$K$ denotes the maximum number of sampling steps,
with $k\in\{1,\ldots,K\}$.
We denote the text embedding by $\boldsymbol{c}$,
the causal student generator by $G_\theta$,
the frozen teacher by $\Phi$, and the fake-score
model by $F_\phi$.
Given initial noise
$\boldsymbol{z}\sim\mathcal{N}(\boldsymbol{0},\mathbf{I})$,
the student output is written as
$\widehat{\boldsymbol{x}}_0^{(k)}
=G_\theta(\boldsymbol{z},\boldsymbol{c},k)$.
For convenience, we define the forward noising operator as
\begin{equation}
\label{eq:forward_operator}
\mathcal{N}_t(\boldsymbol{x}_0,\boldsymbol{\epsilon})
=
\alpha_t\boldsymbol{x}_0
+
\sigma_t\boldsymbol{\epsilon},
\quad
\boldsymbol{\epsilon}
\sim\mathcal{N}(\boldsymbol{0},\mathbf{I}).
\end{equation}

\noindent\textbf{Distribution Matching Distillation.}
In diffusion models, the score function is the gradient
of the log probability density and can be estimated
through noise prediction:
\begin{equation}
\label{eq:score}
\boldsymbol{s}(\boldsymbol{x}_t,t)
=\nabla_{\boldsymbol{x}_t}\log p_t(\boldsymbol{x}_t)
\approx-\frac{\boldsymbol{\epsilon}_{\theta}
(\boldsymbol{x}_t,t)}{\sigma_t},
\end{equation}
where $\boldsymbol{x}_t$ is a noisy latent at timestep
$t$, $\boldsymbol{\epsilon}_{\theta}$ denotes
the predicted noise, and $\sigma_t$ is specified by
the noise schedule. Text conditioning is omitted
for brevity.

DMD minimizes the reverse Kullback--Leibler (KL)
divergence between the noisy student and real-data
distributions. Its generator gradient is approximated by
\begin{equation}
\label{eq:dmd}
\begin{aligned}
\nabla_\theta\mathcal{L}_{\mathrm{DMD}}
&\triangleq
\mathop{\mathbb{E}}\limits_t
\left[
\nabla_\theta D_{\mathrm{KL}}
\left(p_{\mathrm{gen},t}\,\|\,p_{\mathrm{data},t}\right)
\right] \\
&\approx
-\mathop{\mathbb{E}}\limits_{\boldsymbol{z},t,\boldsymbol{\epsilon}}
\left[
\left(
\boldsymbol{s}_{\mathrm{data}}(\boldsymbol{x}_t,t)
-
\boldsymbol{s}_{\mathrm{gen}}(\boldsymbol{x}_t,t)
\right)
\frac{\partial G_\theta}{\partial\theta}
\right],
\end{aligned}
\end{equation}
where $\boldsymbol{s}_{\mathrm{data}}$ and
$\boldsymbol{s}_{\mathrm{gen}}$ are estimated by
the frozen teacher $\Phi$ and the fake-score model
$F_\phi$, respectively. The noisy latent is obtained
as $\boldsymbol{x}_t
=\mathcal{N}_t(\widehat{\boldsymbol{x}}_0^{(k)},
\boldsymbol{\epsilon})$.
The score--Jacobian product denotes contraction over
the latent dimensions, with noise-dependent weighting
omitted for simplicity.

The fake-score model is trained on detached student
samples using the standard denoising objective:
\begin{equation}
\label{eq:fake}
\mathcal{L}_{\mathrm{fake}}
=
\mathop{\mathbb{E}}\limits_{\boldsymbol{z},t,\boldsymbol{\epsilon}}
\left[
\left\|
\boldsymbol{\epsilon}_{F_\phi}
(\operatorname{sg}(\boldsymbol{x}_t),t)
-\boldsymbol{\epsilon}
\right\|_2^2
\right],
\end{equation}
where $\operatorname{sg}(\cdot)$ denotes stop-gradient.
The student generator $G_\theta$ and fake-score model
$F_\phi$ are optimized alternately.

\begin{figure*}[t]
    \centering
    \includegraphics[width=\textwidth]{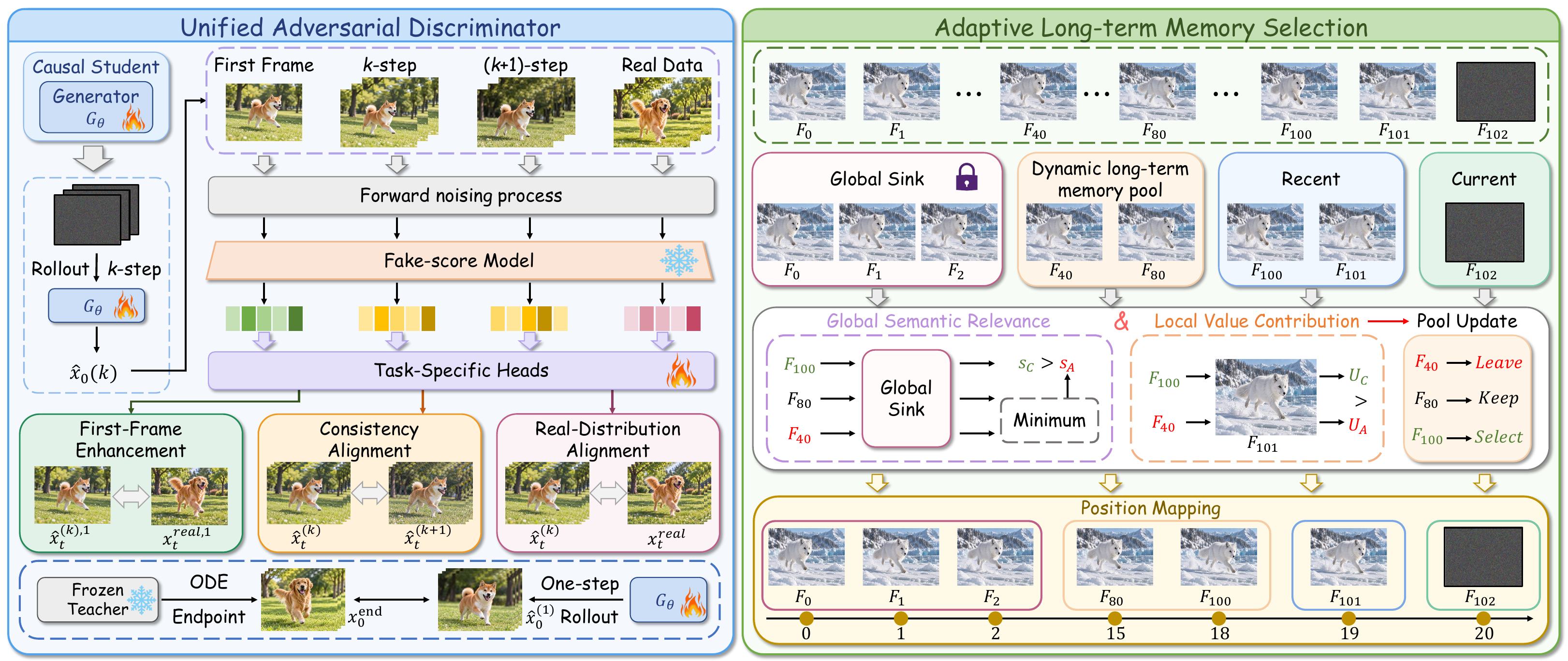}
    \caption{\textbf{Overview of Any Forcing.}
    \textbf{Left:} A unified discriminator jointly performs real-distribution alignment, inter-step consistency alignment, and first-frame enhancement, complemented by teacher ODE endpoint supervision during early training.
    \textbf{Right:} Adaptive long-term memory selection combines a global sink, a dynamic memory pool, and recent local context. Global semantic relevance and local value contribution determine memory admission before generating the next frame, followed by position mapping within the trained RoPE horizon. For better visualization, we use video frames as an example to show the mechanism, where $F_j$ represents the $j$-th frame.}
    \label{fig:method}
\end{figure*}

\section{Methodology}
\label{sec:method}

Any Forcing aims to achieve \emph{any-step generality} and \emph{one-step stability} in autoregressive video generation, with the motivation shown in Fig.~\ref{fig:motivation}.
To address these challenges, we introduce the unified adversarial
discriminator and adaptive long-term memory selection.
Specifically,
as shown in Fig.~\ref{fig:method}, a unified adversarial discriminator jointly performs
real-distribution alignment, inter-step consistency alignment,
and first-frame enhancement, thereby improving \emph{visual quality and sampling flexibility}.
Meanwhile, an adaptive memory selection mechanism
maintains \emph{long-term stability} by selecting historical states
based on global semantic relevance and local value contribution,
while position mapping preserves their temporal relationships.

\subsection{Unified Adversarial Discriminator}
To enable any-step generality while maintaining visual quality, we introduce a unified discriminator (Fig.~\ref{fig:method}, left) that jointly performs real-distribution alignment (RA), inter-step consistency alignment (CA), and first-frame enhancement (FE).

The discriminator uses intermediate features from
the fake-score model $F_\phi$:
\begin{equation}
\label{eq:disc_feature}
\boldsymbol{h}_t
=\mathcal{H}_\phi(\boldsymbol{x}_t,t),
\end{equation}
where $\mathcal{H}_\phi$ denotes its feature extractor.
We denote the unified discriminator by
$D_\psi^{\mathrm{cls}}$, parameterized by $\psi$.
Its shared-backbone classifier produces three task-specific logits:
\begin{equation}
\label{eq:disc_branches}
D_\psi^{\mathrm{cls}}(\boldsymbol{h}_t)
=
\left[
D_\psi^{\mathrm{RA}}(\boldsymbol{h}_t),
D_\psi^{\mathrm{CA}}(\boldsymbol{h}_t),
D_\psi^{\mathrm{FE}}(\boldsymbol{h}_t)
\right],
\end{equation}
where $D_\psi^{\mathrm{RA}}$, $D_\psi^{\mathrm{CA}}$,
and $D_\psi^{\mathrm{FE}}$ denote the scalar logits
for RA, CA, and FE, respectively.
All tasks share the feature extractor, while
task-specific classification heads produce the
corresponding logits.

For each task $a\in\{\mathrm{RA},\mathrm{CA},\mathrm{FE}\}$
and its corresponding task label $\ell_a$,
we define a positive reference $\boldsymbol{u}_0^a$
and a negative student prediction $\boldsymbol{v}_0^a$:
\begin{equation}
\label{eq:task_pairs}
(\boldsymbol{u}_0^a,\boldsymbol{v}_0^a)
=
\begin{cases}
(\boldsymbol{x}_0^{\mathrm{real}},
 \widehat{\boldsymbol{x}}_0^{(k)}),
 & a=\mathrm{RA},\\
(\operatorname{sg}(\widehat{\boldsymbol{x}}_0^{(k+1)}),
 \widehat{\boldsymbol{x}}_0^{(k)}),
 & a=\mathrm{CA},\\
(\boldsymbol{x}_0^{\mathrm{real},1},
 \widehat{\boldsymbol{x}}_0^{(k),1}),
 & a=\mathrm{FE}.
\end{cases}
\end{equation}

RA aligns the full-video distribution with real data.
CA uses the $(k+1)$-step output as the reference for
the $k$-step prediction, where $k<K$ and both rollouts
start from the same initial noise.
FE aligns the first latent frame with real data,
strengthening the initial state that conditions
subsequent autoregressive generation. Following Eq.~\eqref{eq:forward_operator},
we construct the noisy samples as follows:
\begin{equation}
\label{eq:task_noising}
\begin{aligned}
\boldsymbol{u}_t^a
&=\mathcal{N}_t(\boldsymbol{u}_0^a,\boldsymbol{\epsilon}),\\
\boldsymbol{v}_t^a
&=\mathcal{N}_t(\boldsymbol{v}_0^a,\boldsymbol{\epsilon}).
\end{aligned}
\end{equation}

Following~\cite{yang2026towards}, we formulate
all three relativistic adversarial objectives using
$f(u)=-\log(1+e^{-u})$.
The discriminator objective is:
\begin{equation}
\label{eq:unified_d}
\begin{aligned}
\mathcal{L}_D^a
=-\mathbb{E}\Big[
f\Big(&D^a_\psi\big(
\mathcal{H}_\phi(\boldsymbol{u}_t^a,t),\ell_a\big)\\
-&D^a_\psi\big(
\mathcal{H}_\phi(\boldsymbol{v}_t^a,t),\ell_a\big)
\Big)\Big].
\end{aligned}
\end{equation}

The corresponding generator objective reverses
the relative comparison under the same label:
\begin{equation}
\label{eq:unified_g}
\begin{aligned}
\mathcal{L}_G^a
=-\mathbb{E}\Big[
f\Big(&D^a_\psi\big(
\mathcal{H}_\phi(\boldsymbol{v}_t^a,t),\ell_a\big)\\
-&D^a_\psi\big(
\mathcal{H}_\phi(\boldsymbol{u}_t^a,t),\ell_a\big)
\Big)\Big].
\end{aligned}
\end{equation}

\noindent\textbf{Final Objective.}
The discriminator combines the three adversarial
tasks as follows:
\begin{equation}
\label{eq:multi_task_d}
\mathcal{L}_D
=\mathcal{L}_D^{\mathrm{RA}}
+\mathcal{L}_D^{\mathrm{CA}}
+\lambda_{\mathrm{FE}}\mathcal{L}_D^{\mathrm{FE}},
\end{equation}
while the generator adversarial objective is:
\begin{equation}
\label{eq:multi_task_g}
\mathcal{L}_{\mathrm{adv}}
=\mathcal{L}_G^{\mathrm{RA}}
+\mathcal{L}_G^{\mathrm{CA}}
+\lambda_{\mathrm{FE}}\mathcal{L}_G^{\mathrm{FE}}.
\end{equation}

To stabilize early training, we additionally align
the student's one-step prediction with a frozen
teacher ODE endpoint, as shown in
Fig.~\ref{fig:method} (left).
Given the same initial noise $\boldsymbol{z}$ and
text condition $\boldsymbol{c}$, we define:
\begin{equation}
\label{eq:teacher_ode_endpoint}
\boldsymbol{x}_0^{\mathrm{end}}
=\Phi_{\mathrm{ODE}}(\boldsymbol{z},\boldsymbol{c}),
\end{equation}
where $\Phi_{\mathrm{ODE}}$ denotes ODE sampling
with the frozen teacher $\Phi$.
The endpoint alignment loss is:
\begin{equation}
\label{eq:ode_endpoint}
\mathcal{L}_{\mathrm{ODE}}
=\mathbb{E}_{\boldsymbol{z},\boldsymbol{c}}
\left[
\left\|
\widehat{\boldsymbol{x}}_0^{(1)}
-\operatorname{sg}(\boldsymbol{x}_0^{\mathrm{end}})
\right\|_2^2
\right].
\end{equation}

Finally, we combine distribution matching,
unified adversarial supervision, and ODE endpoint
warm-up to optimize the generator:
\begin{equation}
\label{eq:overall_generator}
\mathcal{L}_G
=\mathcal{L}_{\mathrm{DMD}}
+\mathcal{L}_{\mathrm{adv}}
+\lambda_o\mathcal{L}_{\mathrm{ODE}},
\end{equation}
where $\lambda_o=0.05$ is empirically chosen, following the loss-balancing principle in previous work~\cite{yin2024one}.

\subsection{Adaptive Long-Term Memory Selection}
\label{sec:adaptive_memory}

To maintain the visual quality enabled by the unified discriminator over long horizons, we introduce an adaptive long-term memory selection mechanism for stable one-step long-video generation (Fig.~\ref{fig:method}, right). At each attention layer, we maintain three types of historical context: a fixed global sink $\mathcal{S}=\{0,\ldots,S-1\}$, a dynamic long-term memory pool $\mathcal{M}_i$ with capacity $M$, and a recent queue with capacity $N$. Before generating frame $i$, the recent queue contains $[i-N,\ldots,i-1]$, where the oldest frame $C=i-N$ serves as the memory candidate, leaving $\mathcal{R}_i=\{i-N+1,\ldots,i-1\}$, which contains $N-1$ frames, as the recent local context. Since the current frame is initialized from pure Gaussian noise in one-step generation, its query is unreliable for evaluating historical relevance. The related analysis is provided in Supplementary Material~\ref{sec:noisy_query}. We therefore perform memory selection using clean historical states. Candidate $C$ is directly admitted if the pool has available space; otherwise, its admission is determined by two criteria: \emph{Global Semantic Relevance} and \emph{Local Value Contribution}, both calculated from historical states before applying RoPE~\cite{su2024roformer}.

\noindent\textbf{Global Semantic Relevance.}
We measure the relevance between historical frames
$j$ and $l$ using the mean attention logit
over all spatial token pairs and attention heads:
\begin{equation}
\label{eq:summary_response}
a(j,l)
=\frac{1}{HP^2\sqrt{d}}
\sum_{h=1}^{H}\sum_{p=1}^{P}\sum_{u=1}^{P}
(\boldsymbol{q}_{j,p}^{h})^\top
\boldsymbol{k}_{l,u}^{h},
\end{equation}
where $P$ is the number of spatial tokens,
$H$ is the number of attention heads, and $d$
is the head dimension.

Global relevance is measured by averaging
the responses to all sink frames:
\begin{equation}
\label{eq:global_relevance}
s_j=\frac{1}{S}\sum_{l\in\mathcal{S}}a(j,l),
\quad
A=\operatorname*{arg\,min}_{j\in\mathcal{M}_i}s_j.
\end{equation}

Candidate $C$ undergoes a further local evaluation only if $s_C>s_A$, indicating higher global relevance than the least relevant memory entry.

\noindent\textbf{Local Value Contribution.}
After satisfying the global semantic relevance criterion, candidate $C$ is compared with memory entry $A$ based on their relevance to the recent context. Their responses are jointly normalized for each recent frame $j\in \mathcal{R}_i$, and the resulting weights are averaged:
\begin{equation}
\label{eq:local_competition}
\begin{aligned}
&(p_A^{(j)},p_C^{(j)})
=\operatorname{softmax}\big([a(j,A),a(j,C)]\big),\\
&U_v=\frac{1}{N-1}
\sum_{j\in\mathcal{R}_i}p_v^{(j)},
\quad v\in\{A,C\}.
\end{aligned}
\end{equation}

Since both entries compete under the same
normalization, $U_C>U_A$ indicates a stronger
relative preference for $C$ within the local context.
For a full pool, we update
\begin{equation}
\label{eq:memory_update}
\mathcal{M}_i^{*}=
\begin{cases}
(\mathcal{M}_i\setminus\{A\})\cup\{C\},
&s_C>s_A\ \land\ U_C>U_A,\\
\mathcal{M}_i,&\text{otherwise}.
\end{cases}
\end{equation}

The current frame is generated using global sink
$\mathcal{S}$, updated memory $\mathcal{M}_i^{*}$,
and the remaining recent frames in
$\mathcal{R}_i$.

\noindent\textbf{Position Mapping.}
To maintain long-video generation within the trained RoPE horizon $L$, we remap historical frames while preserving their temporal order. The global sink retains its original positions $\{0,\ldots,S-1\}$, recent frames are assigned $\{L-N,\ldots,L-2\}$, and the current frame uses $L-1$, while long-term memory is mapped to $\{S,\ldots,L-N-1\}$.

For the long-term memory pool $\mathcal{M}_i^{*}$ containing $n_i$ frames, their original temporal positions are denoted by $m_0<\cdots<m_{n_i-1}$. For a nonempty long-term memory pool, the source and target upper bounds are defined as:
\begin{equation}
\begin{aligned}
b_i&=
\begin{cases}
i-N,&N>1,\\
m_{n_i-1},&N=1,
\end{cases}\\
h&=L-N-1.
\end{aligned}
\end{equation}

Each original temporal position $m_p$, where $p\in\{0,\ldots,n_i-1\}$, is linearly mapped to a new RoPE position $\pi_i(m_p)$:
\begin{equation}
\label{eq:memory_remapping}
\pi_i(m_p)=\operatorname{round}\left[
S+\frac{m_p-S}{\max(b_i-S,1)}(h-S)
\right].
\end{equation}

Under $S+M+N\leq L$, the remapped positions remain within the trained horizon while preserving temporal order. The adaptive memory selection mechanism is activated during the generation of long-videos.

%% file: sec/4_details.tex
\begin{table*}[t]
\centering
\caption{Comparison on VBench for 5-second video generation at 480p resolution.
For the few-step setting, the left and right groups use four and two denoising steps, respectively.
For the one-step setting, $1^{*}$ and $1^{\dagger}$ use four and two denoising steps for the first latent frame, respectively, followed by one step for subsequent frames.
Best and second-best results within each sampling group are \textbf{bold} and \underline{underlined}, respectively.
Sem.: Semantic. We report three representative quality and semantic metrics here: IQ: Imaging Quality; AQ: Aesthetic Quality; OC: Object Class. Detailed results can be found in Supplementary Material~\ref{more_res}.}
\label{tab:vbench_comparison}
\setlength{\tabcolsep}{3.7pt}
\renewcommand{\arraystretch}{1.0}
\small
\begin{tabular}{
l!{\vrule width 0.4pt}
ccccccc!{\vrule width 0.4pt}
ccccccc
}
\toprule
Method
& NFE & Total$\uparrow$ & Quality$\uparrow$ & Sem.$\uparrow$
& IQ$\uparrow$ & AQ$\uparrow$ & OC$\uparrow$
& NFE & Total$\uparrow$ & Quality$\uparrow$ & Sem.$\uparrow$
& IQ$\uparrow$ & AQ$\uparrow$ & OC$\uparrow$ \\
\midrule
\rowcolor{gray!35}
\textit{Base model} & & & & & & & & & & & & & & \\
Wan 2.1-1.3B~\cite{wan2025}
& 50 & 83.99 & 84.82 & 80.66 & 67.52 & 65.70 & 94.56
& 20 & 83.63 & 84.52 & 80.06 & 65.71 & 65.19 & 93.81 \\
\midrule
\rowcolor{gray!35}
\textit{Few steps} & & & & & & & & & & & & & & \\
CausVid~\cite{yin2025causvid}
& 4 & 83.05 & 84.14 & 78.70 & 68.44 & 66.04 & 93.35
& 2 & 81.03 & 82.26 & 76.11 & 63.25 & 62.55 & 90.22 \\
rCM~\cite{zheng2026large}
& 4 & 83.11 & 84.48 & 77.65 & 66.30 & 63.33 & 92.96
& 2 & 81.06 & 83.23 & 72.34 & 59.84 & 62.04 & 90.35 \\
Self Forcing~\cite{huang2026self}
& 4 & 84.20 & 84.88 & \textbf{81.48} & 69.44 & 66.68 & 94.37
& 2 & 81.79 & 83.13 & 76.40 & 66.02 & 63.39 & 90.40 \\
Causal Forcing~\cite{zhu2026causal}
& 4 & \underline{85.08} & \textbf{86.19} & 80.64 & \underline{70.37} & \textbf{67.53} & \underline{95.62}
& 2 & \underline{84.03} & \underline{85.18} & \underline{79.40} & \underline{68.49} & \underline{66.26} & \underline{95.22} \\
\rowcolor{gray!20}
\textbf{Any Forcing} (Ours)
& 4 & \textbf{85.09} & \underline{86.18} & \underline{80.73} & \textbf{71.68} & \underline{67.41} & \textbf{96.28}
& 2 & \textbf{84.89} & \textbf{86.08} & \textbf{80.12} & \textbf{70.81} & \textbf{66.53} & \textbf{96.22} \\
\midrule
\rowcolor{gray!35}
\textit{One step} & & & & & & & & & & & & & & \\
Self Forcing~\cite{huang2026self}
& $1^{*}$ & 83.08 & 83.73 & 80.50 & 66.46 & 64.66 & 91.52
& $1^{\dagger}$ & 82.25 & 83.16 & 78.62 & 66.44 & 63.42 & 91.04 \\
Causal Forcing~\cite{zhu2026causal}
& $1^{*}$ & 83.06 & 83.75 & 80.28 & 64.01 & 63.30 & 92.69
& $1^{\dagger}$ & 81.87 & 82.66 & 78.73 & 66.12 & \underline{64.71} & 92.28 \\
ASD~\cite{yang2026towards}
& $1^{*}$ & 82.74 & 83.08 & \textbf{81.35} & \underline{68.98} & \underline{66.00} & 94.40
& $1^{\dagger}$ & 81.70 & 82.29 & 79.32 & \underline{66.70} & 64.59 & \underline{96.19} \\
One Forcing~\cite{feng2026oneforcing}
& $1^{*}$ & \underline{84.35} & \underline{85.52} & 79.65 & 66.67 & 64.93 & \underline{95.13}
& $1^{\dagger}$ & \underline{84.01} & \underline{85.11} & \underline{79.59} & 65.61 & 64.53 & 95.93 \\
\rowcolor{gray!20}
\textbf{Any Forcing} (Ours)
& $1^{*}$ & \textbf{85.06} & \textbf{86.19} & \underline{80.54} & \textbf{70.35} & \textbf{66.31} & \textbf{95.25}
& $1^{\dagger}$ & \textbf{84.79} & \textbf{85.83} & \textbf{80.64} & \textbf{69.72} & \textbf{65.78} & \textbf{96.57} \\
\bottomrule
\end{tabular}
\end{table*}

\begin{table*}[t]
\centering
\caption{Comparison on VBench for 30- and 60-second video generation at 480p resolution.
$1^{*}$ and $1^{\dagger}$ use four and two sampling steps for the first latent frame, respectively, followed by one step for subsequent frames.
Best and second-best results within each one-step sampling group and video length are \textbf{bold} and \underline{underlined}, respectively.}
\label{tab:long_video_comparison}
\setlength{\tabcolsep}{3.7pt}
\renewcommand{\arraystretch}{1.0}
\small
\begin{tabular}{
l!{\vrule width 0.4pt}
ccccccc!{\vrule width 0.4pt}
ccccccc
}
\toprule
\multirow{2}{*}{Method}
& \multicolumn{7}{c!{\vrule width 0.4pt}}%
  {\textit{30-second video generation}}
& \multicolumn{7}{c}%
  {\textit{60-second video generation}} \\
\cmidrule(lr){2-8}\cmidrule(lr){9-15}
& NFE & Total$\uparrow$ & Quality$\uparrow$ & Sem.$\uparrow$
& IQ$\uparrow$ & AQ$\uparrow$ & OC$\uparrow$
& NFE & Total$\uparrow$ & Quality$\uparrow$ & Sem.$\uparrow$
& IQ$\uparrow$ & AQ$\uparrow$ & OC$\uparrow$ \\
\midrule
\rowcolor{gray!35}
\textit{Few steps} & & & & & & & & & & & & & & \\
CausVid~\cite{yin2025causvid}
& 4 & 78.54 & 81.08 & 68.38 & 60.30 & 55.42 & 81.05
& 4 & 74.92 & 79.29 & 57.42 & 56.80 & 50.78 & 65.50 \\
LongLive~\cite{yang2026longlive}
& 4 & 82.81 & 83.34 & 80.69 & 68.73 & 63.90 & 95.12
& 4 & 82.73 & 83.24 & 80.69 & 68.57 & 64.16 & 94.43 \\
\midrule
\rowcolor{gray!35}
\textit{One step} & & & & & & & & & & & & & & \\
Infinity RoPE~\cite{yesiltepe2026infinity}
& $1^{*}$ & \underline{84.51} & \textbf{85.83} & 79.22 & \textbf{70.52} & 62.77 & \underline{94.88}
& $1^{*}$ & 84.12 & \underline{85.57} & 78.31 & \textbf{70.80} & 61.46 & 93.41 \\
Deep Forcing~\cite{yi2025deep}
& $1^{*}$ & 84.33 & 85.43 & \underline{79.91} & 69.97 & \underline{63.75} & 94.66
& $1^{*}$ & \underline{84.26} & 85.37 & \textbf{79.79} & 70.00 & \underline{63.24} & \underline{94.53} \\
Relax Forcing~\cite{zhao2026relax}
& $1^{*}$ & 84.09 & 85.27 & 79.40 & 69.06 & 63.12 & 93.11
& $1^{*}$ & 84.00 & 85.29 & 78.85 & 69.13 & 62.80 & 93.09 \\
\rowcolor{gray!20}
\textbf{Any Forcing} (Ours)
& $1^{*}$ & \textbf{84.56} & \underline{85.64} & \textbf{80.22} & \underline{70.14} & \textbf{64.06} & \textbf{95.73}
& $1^{*}$ & \textbf{84.46} & \textbf{85.69} & \underline{79.53} & \underline{70.15} & \textbf{63.55} & \textbf{95.87} \\
\hdashline
Infinity RoPE~\cite{yesiltepe2026infinity}
& $1^{\dagger}$ & \underline{84.13} & \underline{85.56} & 78.42 & \textbf{70.30} & 62.04 & 93.29
& $1^{\dagger}$ & 83.81 & \underline{85.31} & 77.81 & \textbf{70.26} & 60.98 & 92.67 \\
Deep Forcing~\cite{yi2025deep}
& $1^{\dagger}$ & 84.06 & 85.30 & \underline{79.10} & 69.63 & \underline{63.42} & \underline{94.61}
& $1^{\dagger}$ & \underline{83.90} & 85.21 & \underline{78.63} & 69.50 & \underline{62.60} & \underline{94.52} \\
Relax Forcing~\cite{zhao2026relax}
& $1^{\dagger}$ & 83.76 & 85.10 & 78.41 & 68.73 & 62.65 & 92.42
& $1^{\dagger}$ & 83.67 & 85.10 & 77.92 & 68.64 & 62.21 & 92.56 \\
\rowcolor{gray!20}
\textbf{Any Forcing} (Ours)
& $1^{\dagger}$ & \textbf{84.53} & \textbf{85.73} & \textbf{79.71} & \underline{69.98} & \textbf{63.59} & \textbf{94.80}
& $1^{\dagger}$ & \textbf{84.23} & \textbf{85.41} & \textbf{79.51} & \underline{69.75} & \textbf{63.13} & \textbf{94.77} \\
\bottomrule
\end{tabular}
\end{table*}

\begin{figure*}[t]
    \centering
    \includegraphics[width=\textwidth]{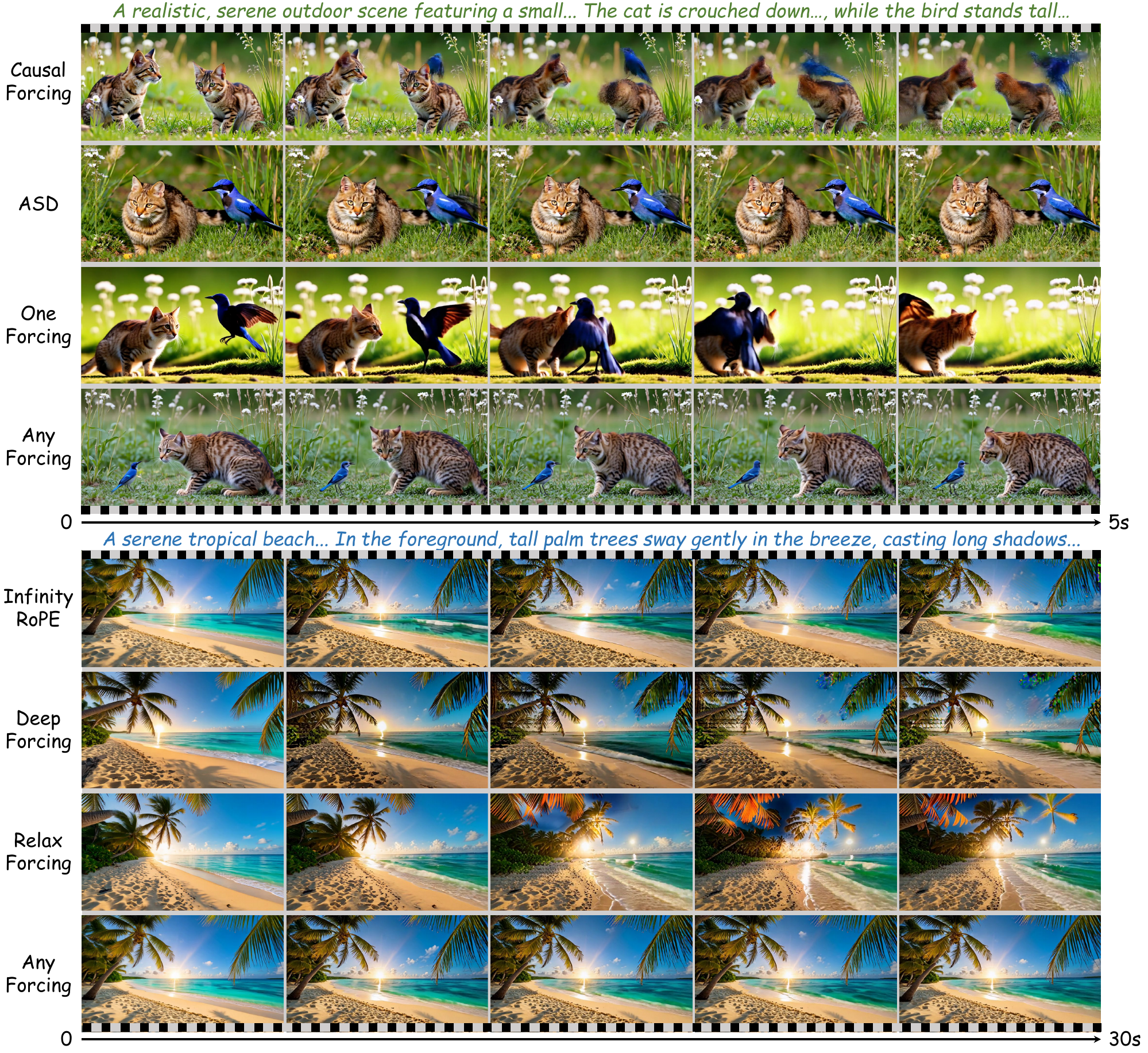}
    \caption{Visual comparisons of one-step short-video
    (top) and long-video (bottom) generation.
    Any Forcing achieves high-quality video generation while preserving stable subject appearance and scene structure over time. All methods adopt the $1^{*}$ sampling configuration. More visualization results are available in Supplementary Material~\ref{more_res}.}
    \label{fig:vis}
\end{figure*}
\section{Experiments}
\subsection{Experimental Setup}
\noindent\textbf{Implementation Details.}
We follow the training setups of Causal
Forcing~\cite{zhu2026causal} and ASD~\cite{yang2026towards},
initializing the student from the ODE-distilled checkpoint
released by Causal Forcing.
Training uses its real videos and teacher ODE trajectories,
with at most four sampling steps.
Our model is trained for 1,000 steps.
We apply ODE endpoint supervision during the first 200 steps
and, following DMD~\cite{yin2024one}, perform five fake-score
updates per generator update.
For long-video inference, we retain three global sink frames,
two long-term memory frames, and one recent context frame.
Experiments are conducted on 16 NVIDIA A100 GPUs. 
We evaluate 5-second videos using VBench~\cite{huang2024vbench}
and 30- and 60-second videos using VBench++~\cite{huang2025vbench++},
all at 480p resolution.
For robust and statistically reliable evaluation, we perform five independent samplings per prompt. Main comparisons cover all 946 prompts, while long-video ablations and hyperparameter analyses use a fixed subset of 128 prompts for efficiency, following~\cite{yi2025deep}. More implementation details are available in Supplementary Material~\ref{detail}.

\noindent\textbf{Baselines.}
For short-video generation, we compare with the base model
Wan2.1~\cite{wan2025} and representative few-step and
one-step-capable methods, including
CausVid~\cite{yin2025causvid},
rCM~\cite{zheng2026large},
Self Forcing~\cite{huang2026self},
Causal Forcing~\cite{zhu2026causal},
ASD~\cite{yang2026towards}, and
One Forcing~\cite{feng2026oneforcing}.
For long-video generation, we include CausVid and
LongLive~\cite{yang2026longlive} as few-step baselines.
For the one-step comparison, we apply the training-free
strategies Infinity RoPE~\cite{yesiltepe2026infinity},
Deep Forcing~\cite{yi2025deep}, and
Relax Forcing~\cite{zhao2026relax} to our generator
trained with the unified adversarial discriminator,
which achieves the highest total score among the evaluated short-video models in one-step.
Using the same generator enables a fair comparison.

%% file: sec/5_results.tex
\subsection{Main Results}

\noindent\textbf{Video Generation.}
As shown in Tables~\ref{tab:vbench_comparison}
and~\ref{tab:long_video_comparison}, Any Forcing achieves
the highest total scores across the evaluated short-video
sampling configurations and both one-step configurations
for long videos.
For short videos, reducing sampling from four to two steps
decreases its total score by only $0.24\%$ ($85.09 \rightarrow 84.89$),
compared with $1.23\%$ ($85.08 \rightarrow 84.03$) for Causal Forcing
and $2.86\%$ ($84.20 \rightarrow 81.79$) for Self Forcing,
demonstrating stronger robustness to reduced sampling budgets.
With first-frame sampling fixed at four or two steps,
reducing subsequent-frame sampling to one step yields
total-score decreases of only $0.04\%$ ($85.09 \rightarrow 85.06$)
and $0.12\%$ ($84.89 \rightarrow 84.79$), respectively,
whereas the corresponding decreases for Causal Forcing are
$2.37\%$ ($85.08 \rightarrow 83.06$)
and $2.57\%$ ($84.03 \rightarrow 81.87$).
In particular, the $1^{*}$ configuration closely matches
full four-step sampling, providing an effective balance
between generation quality and sampling cost.
The $1^{\dagger}$ configuration also achieves the highest
quality and semantic scores among the compared one-step methods.
For both 30- and 60-second videos, Any Forcing achieves the best or second-best results across all reported metrics under both first-frame sampling settings, with the highest total, aesthetic-quality, and object-class scores.
When extending generation from 30 to 60 seconds,
its total score decreases by $0.12\%$
($84.56 \rightarrow 84.46$) under $1^{*}$ and
$0.35\%$ ($84.53 \rightarrow 84.23$) under $1^{\dagger}$.
Across the two video lengths, its aesthetic-quality score
changes from $64.06$ to $63.55$ under $1^{*}$ and
from $63.59$ to $63.13$ under $1^{\dagger}$,
while the corresponding object-class scores remain
close at $95.73$ and $95.87$, and $94.80$ and $94.77$,
respectively.
These results indicate that Any Forcing enables stable long-horizon video generation.
The visual comparisons in Fig.~\ref{fig:vis} further
illustrate the visual quality and content stability
of Any Forcing.
In the short-video example, Any Forcing preserves distinct
cat and bird identities with consistent appearances,
while some baselines exhibit subject confusion or blurring.
In the 30-second video sequence, our method maintains
coherent palm-tree structures, shoreline geometry, and
consistent illumination, whereas Relax Forcing exhibits
content collapse and color drift, and other baselines also
suffer from blurring and content collapse.
Together, these results support the any-step generality
of Any Forcing and its ability to maintain high-quality,
stable one-step generation over long sequences.

\begin{figure}[t]
    \centering
    \includegraphics[width=\linewidth]{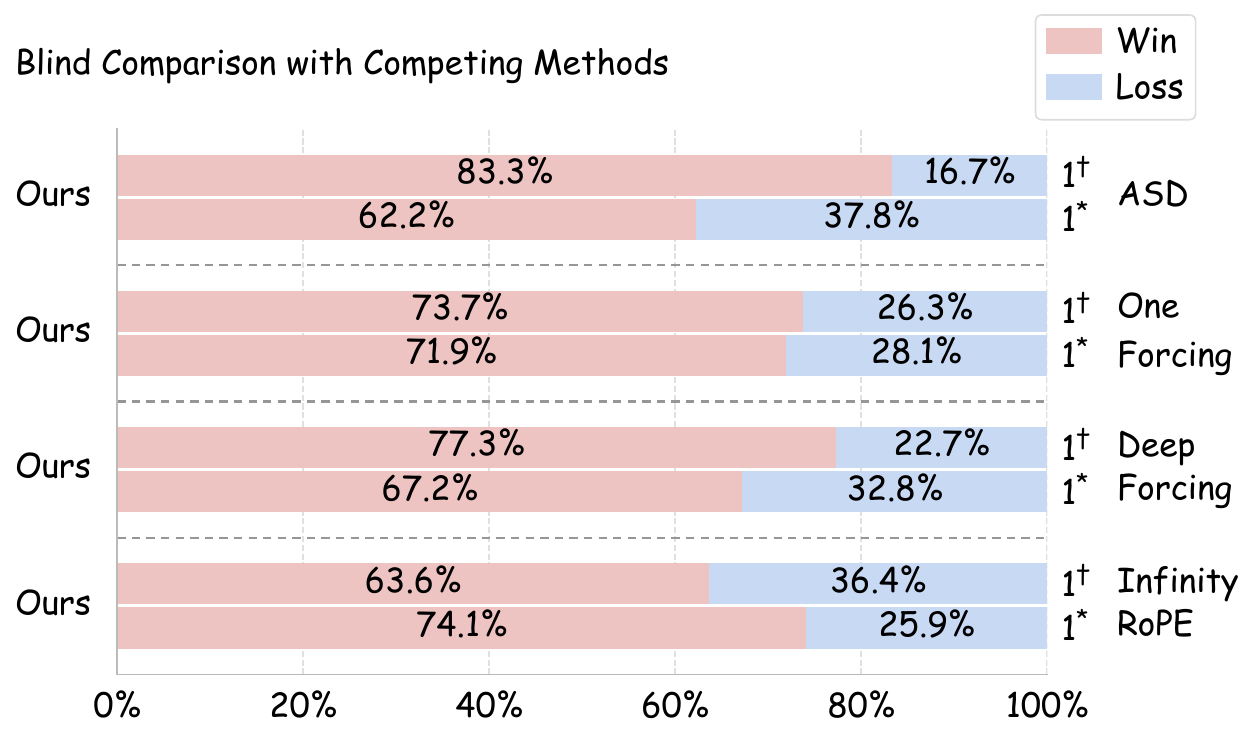}
    \caption{Human preference results.
    The corresponding sampling settings are indicated to the right of the bars.
    We report the win rate of Any Forcing against others in blind pairwise comparisons.}
    \label{fig:user_study}
\end{figure}

\noindent\textbf{Human Preference.}
We conduct a blind pairwise user study with 48 evaluators. Participants compare videos generated from the same prompt and presented in randomized left--right order, selecting the preferred video based on visual quality and semantic alignment. As shown in Fig.~\ref{fig:user_study}, Any Forcing is
consistently preferred over all compared methods.
For short videos, its preference rates against ASD and One Forcing are 62.2\% and 71.9\% under $1^{*}$, and 83.3\% and 73.7\% under $1^{\dagger}$.
For long videos, the corresponding rates against
Deep Forcing and Infinity RoPE are 67.2\% and 74.1\% under $1^{*}$, and 77.3\% and 63.6\% under $1^{\dagger}$. These results demonstrate the advantages of our method in \emph{achieving high-quality and stable one-step video generation}.  Details are available in Supplementary Material~\ref{sec:human_preference_details}.

\begin{figure}[t]
    \centering
    \includegraphics[width=\linewidth]{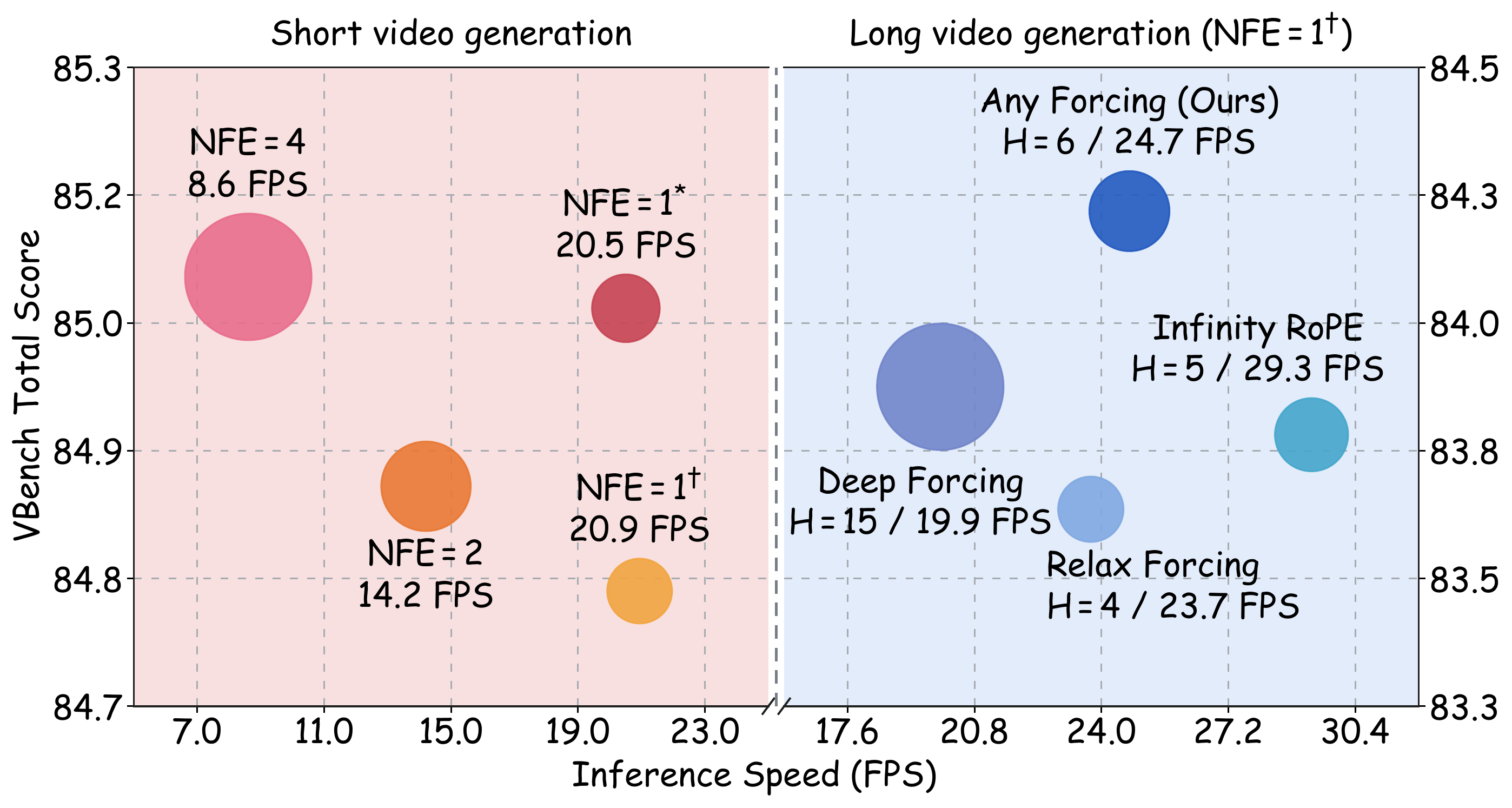}
    \caption{Performance analysis of short (left) and long (right)
    video generation on a single H100 GPU.
    $1^{*}$ and $1^{\dagger}$ use four and two first-frame
    steps, respectively, followed by one step per frame.
    The bubble size indicates the number of sampling steps (left) and the historical-frame budget $H$, excluding the current frame (right).}
    \label{fig:speed_quality}
\end{figure}

\noindent\textbf{Performance Analysis.}
Fig.~\ref{fig:speed_quality} examines the quality--efficiency trade-offs of short- and long-video generation.
For short videos, we compare sampling schedules within Any Forcing, since methods sharing the same architecture and sampling schedule incur the same inference costs.
Comparing the $1^{*}$ and $1^{\dagger}$ settings, reducing first-frame steps from four to two lowers the total score from 85.06 to 84.79, highlighting the importance of initial-context quality.
More steps for the first frame and one step thereafter balances quality and efficiency.
For one-step long-video generation, Any Forcing achieves the highest total score with competitive throughput.
Our method achieves \emph{20.9 FPS} and \emph{24.7 FPS} for short- and long-video generation, respectively.

\begin{table}[t]
\centering
\caption{Ablation study of the proposed components on VBench.
The left and right groups correspond to $1^{*}$ and $1^{\dagger}$ sampling, respectively.
Best and second-best results within each sampling group are \textbf{bold} and \underline{underlined}, respectively.}
\label{tab:ablation}

\setlength{\tabcolsep}{1.5pt}
\renewcommand{\arraystretch}{1.0}
\small

\begin{tabular}{
l!{\vrule width 0.4pt}
ccc!{\vrule width 0.4pt}
ccc
}
\toprule
Method
& Total & Quality & Semantic
& Total & Quality & Semantic \\
\midrule

\rowcolor{gray!20}
\textbf{Any Forcing} 
& \textbf{85.06} & \textbf{86.19} & \underline{80.54}
& \textbf{84.79} & \textbf{85.83} & \textbf{80.64} \\

w/o CA
& 84.66 & 85.67 & \textbf{80.62}
& \underline{84.14} & 85.11 & \underline{80.25} \\

w/o RA
& 83.78 & 85.19 & 78.15
& 83.86 & \underline{85.12} & 78.85 \\

w/o FE
& 83.52 & 84.86 & 78.15
& 81.86 & 83.22 & 76.45 \\

w/o ODE
& \underline{84.67} & \underline{85.83} & 80.00
& 83.10 & 84.17 & 78.84 \\

\bottomrule
\end{tabular}
\end{table}

\subsection{Ablation Studies}

\noindent\textbf{Unified Adversarial Discriminator.}
Table~\ref{tab:ablation} demonstrates the complementary
contributions of the three adversarial tasks under both
sampling settings.
Removing consistency alignment (CA) or real-distribution
alignment (RA) reduces overall performance, highlighting
the importance of aligning sampling-step outputs and
the real data distribution.
First-frame enhancement (FE) has the largest impact:
its removal lowers the total score from 85.06 to 83.52
under $1^{*}$ and from 84.79 to 81.86 under $1^{\dagger}$.
The larger decline under $1^{\dagger}$ indicates the
importance of first-frame quality when fewer initial
sampling steps are used.
Although removing CA slightly improves the semantic score
under $1^{*}$, the complete model achieves the highest
total and quality scores under both settings.
Removing ODE endpoint supervision also degrades performance, supporting its role in stabilizing early training. Detailed analysis of its effect on training stability is provided in Supplementary Material~\ref{ode_warm}.

\noindent\textbf{Adaptive Long-Term Memory Selection.}
Table~\ref{tab:memory_ablation} demonstrates the complementary roles of global semantic relevance (GSR), local value contribution (LVC), and position mapping (PM).
Removing GSR lowers the object-class score from 93.50 to 92.67, while removing LVC reduces the scene score from 59.70 to 59.15, supporting their respective roles in retaining globally relevant and locally useful historical states.
Replacing position mapping (w/o PM) with direct concatenation of the selected historical states before the recent context decreases imaging quality from 71.71 to 70.90 and degrades several semantic metrics, highlighting the importance of preserving temporal relationships within the compressed cache.
The complete mechanism achieves the best performance across most evaluation dimensions.
Further analyses of positional mapping and memory selection are available in Supplementary Material~\ref{sec:memory_analysis}.

\subsection{Hyperparameter Analysis}

\noindent\textbf{First-Frame Enhancement Weight.}
Table~\ref{tab:hyperparameter} evaluates the sensitivity
to $\lambda_{\mathrm{FE}}$.
Setting $\lambda_{\mathrm{FE}}=0.05$ achieves the best
performance across all reported metrics and approximately
matches the proportion of the first frame in the latent sequence (1/21).
A smaller weight may provide insufficient first-frame
supervision, limiting the quality of the initial context
on which subsequent autoregressive generation depends.
Conversely, a larger weight may overemphasize first-frame
quality at the expense of the full-sequence objectives,
potentially compromising temporal continuity.
The results suggest that a moderate weight better balances
first-frame enhancement with overall video quality and
semantic alignment.
We therefore adopt $\lambda_{\mathrm{FE}}=0.05$ by default.

\begin{table}[t]
\centering
\caption{Ablation study of adaptive long-term memory selection
on VBench. GSR: Global Semantic Relevance;
LVC: Local Value Contribution; PM: Position Mapping.
Best and second-best results are \textbf{bold} and \underline{underlined}, respectively.}
\label{tab:memory_ablation}
\setlength{\tabcolsep}{1.5pt}
\renewcommand{\arraystretch}{1.0}
\small
\begin{tabular}{l*{8}{c}}
\toprule
Method & IQ & AQ & OC & MO & SR & TF & Scene & TS \\
\midrule
\rowcolor{gray!20}
\textbf{Any Forcing}
& \textbf{71.71} & \textbf{59.38}
& \textbf{93.50} & \textbf{82.08}
& 89.14 & \underline{99.34}
& \textbf{59.70} & \textbf{23.28} \\
w/o GSR
& 71.68 & 59.33
& 92.67 & 81.55
& \underline{89.55} & \textbf{99.37}
& \underline{59.36} & \underline{23.20} \\
w/o LVC
& \underline{71.69} & \underline{59.35}
& \underline{93.27} & \underline{81.72}
& \textbf{89.73} & 99.33
& 59.15 & 23.09 \\
w/o PM
& 70.90 & 58.99
& 92.45 & 81.44
& 89.51 & 99.27
& 59.05 & 22.95 \\
\bottomrule
\end{tabular}
\end{table}

\begin{table}[t]
\centering
\caption{Sensitivity analysis of the first-frame enhancement
weight $\lambda_{\mathrm{FE}}$ on VBench.
The left and right groups correspond to $1^{*}$ and $1^{\dagger}$ sampling, respectively.
Best and second-best results within each sampling group are \textbf{bold} and \underline{underlined}, respectively.}
\label{tab:hyperparameter}

\setlength{\tabcolsep}{3.5pt}
\renewcommand{\arraystretch}{1.0}
\small

\begin{tabular}{
c!{\vrule width 0.4pt}
ccc!{\vrule width 0.4pt}
ccc
}
\toprule
$\lambda_{\mathrm{FE}}$
& Total & Quality & Semantic
& Total & Quality & Semantic \\
\midrule

0.01
& 83.46 & \underline{84.95} & 77.49
& 81.59 & \underline{82.94} & 76.23 \\

\rowcolor{gray!20}
\textbf{0.05}
& \textbf{85.06} & \textbf{86.19} & \textbf{80.54}
& \textbf{84.79} & \textbf{85.83} & \textbf{80.64} \\

0.10
& \underline{83.54} & 84.88 & \underline{78.16}
& \underline{81.64} & 82.78 & \underline{77.08} \\

\bottomrule
\end{tabular}
\end{table}

\begin{figure}[t]
    \centering
    \includegraphics[width=\linewidth]{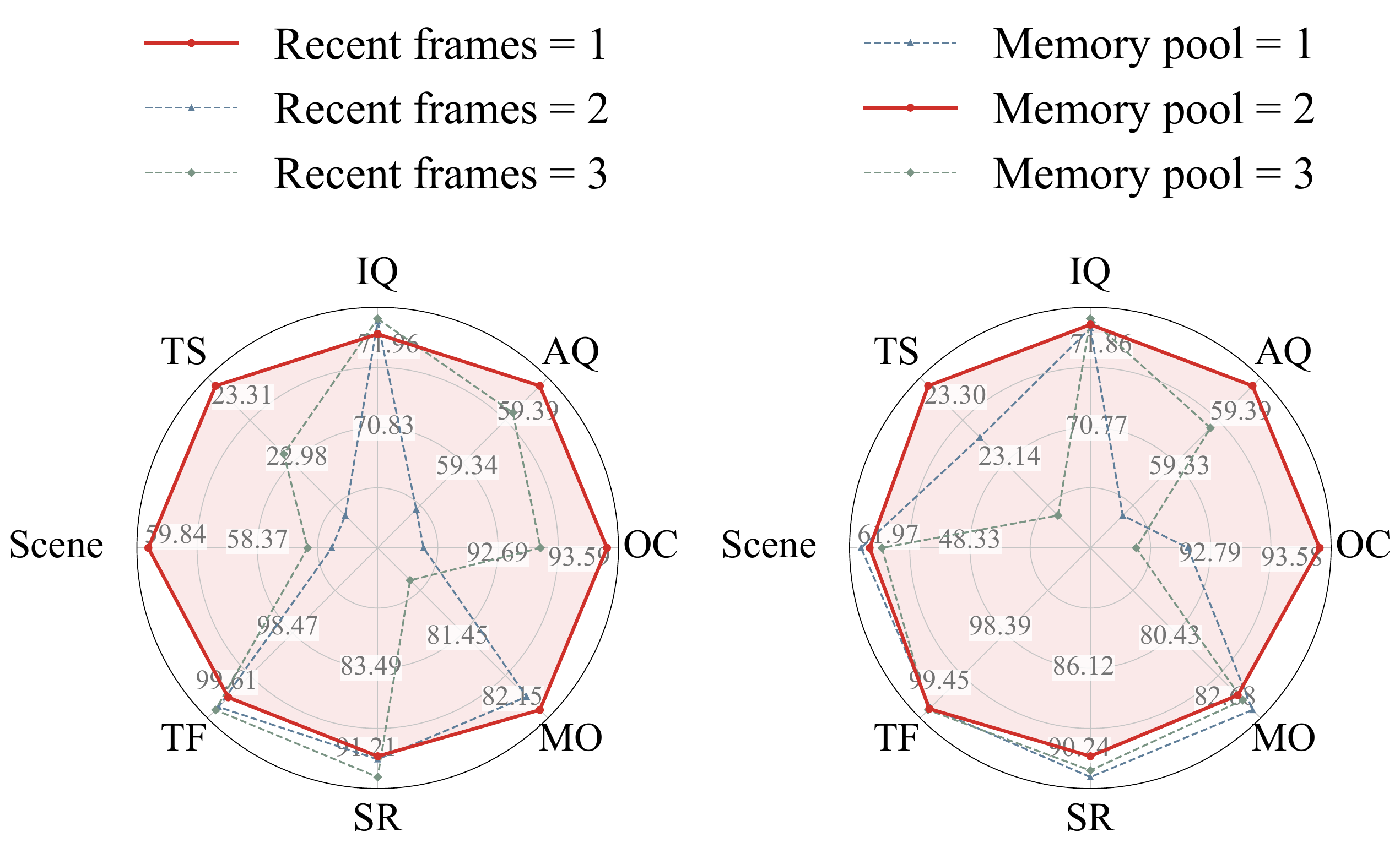}
    \caption{Sensitivity analysis of recent-context length (left)
    and long-term memory capacity (right). Recent frames denote the number of retained local-context frames after candidate removal.}
    \label{fig:context_hyperparameter}
\end{figure}

\noindent\textbf{Memory Pool and Recent Context Analysis.}
Fig.~\ref{fig:context_hyperparameter} shows the effects
of recent-context length and memory-pool capacity.
A longer recent context may propagate errors from
more consecutive generated frames, increasing the risk of error accumulation and local drift
in subsequent generation.
A smaller memory pool limits the amount of historical information retained, potentially weakening
long-term contextual support.
Conversely, a larger pool may introduce redundant
or outdated states that compete with relevant context
and interfere with the current generation.
The observed metric trade-offs suggest that more
cached history does not necessarily yield better results.
We therefore retain one recent frame and two memory
frames.

%% file: sec/6_conclusion.tex
\section{Conclusion}

In this paper, we propose Any Forcing, a novel method for \emph{any-step generality and one-step stability} in autoregressive video generation. Our unified adversarial discriminator jointly performs real-distribution alignment, inter-step consistency alignment, and first-frame enhancement, enabling flexible sampling while maintaining high-quality one-step generation. Meanwhile, an adaptive long-term memory selection mechanism retains globally relevant and locally useful historical states to improve long-horizon stability. Extensive experiments on short- and long-video generation demonstrate consistent advantages over the compared methods. A remaining limitation is that attention-based memory selection cannot capture complex scene dynamics over long sequences. Future work will explore more stable memory strategies under complex motion and scene dynamics.

%% file: sec/supplementary.tex
\newpage
\appendix
\onecolumn

\begin{center}
{\bfseries\LARGE Supplementary Material}\\[0.05in]
\end{center}

\section{Details of Implementations}
\label{detail}

\noindent\textbf{Model Architecture and Initialization.}
Our causal student generator and fake-score model use the
Wan2.1-T2V-1.3B backbone, while the frozen real-score teacher
uses Wan2.1-T2V-14B.
The student is initialized from the framewise ODE-initialization
checkpoint released by Causal Forcing.
Training sequences contain 21 latent frames, each with
16 channels and a spatial resolution of $60\times104$,
corresponding to 81 decoded frames at 480p resolution.

\noindent\textbf{Training Data and Sampling.}
We use the real video latents, text conditions, and teacher
ODE trajectories provided by Causal Forcing.
The maximum sampling budget is four steps, using the raw
timestep schedule $\{1000,750,500,250\}$.
We apply a timestep shift of 5 and randomly select the
sampling budget during training.
The fake-score model is trained with the flow-matching
denoising objective, and the real-score teacher uses
classifier-free guidance with a scale of 3. 

\noindent\textbf{Optimization.}
We use AdamW with $\beta_1=0$ and $\beta_2=0.999$.
The learning rates are $2\times10^{-6}$ for the generator
and $4\times10^{-7}$ for both the fake-score model and
discriminator heads.
Each GPU processes one sequence per micro-batch, with four
gradient accumulation steps, giving an effective batch
size of 64 sequences across 16 GPUs.
Five fake-score updates are performed per generator update.
Each fake-score update is accompanied by a separate
discriminator-head update.
The discriminator backbone receives updates through the
fake-score objective, while adversarial discriminator
updates optimize its heads.
We use mixed precision and gradient checkpointing.
Generator EMA starts at step 200 with a decay of 0.99,
and EMA weights are used for evaluation. Further implementation details are provided in Table~\ref{tab:implementation_details}

\begin{table*}[h]
\centering
\caption{Implementation settings of Any Forcing.}
\label{tab:implementation_details}
\setlength{\tabcolsep}{6pt}
\renewcommand{\arraystretch}{1.15}
\small
\begin{tabular}{p{0.30\textwidth}p{0.64\textwidth}}
\toprule
Hyperparameter & Setting \\
\midrule
Generator initialization
& ODE-initialization causal model from Causal Forcing \\
Generator / real score / fake score
& Wan2.1-T2V-1.3B / Wan2.1-T2V-14B / Wan2.1-T2V-1.3B \\
Objective
& Flow denoising for fake score;
DMD + multi-task GAN + ODE endpoint supervision for generator \\
Training frames
& 21 latent frames; $16\times60\times104$ per frame \\
Frames per autoregressive block & 1 \\
Training sampling steps & Randomly selected from $\{1,2,3,4\}$ \\
Guidance scale & Real-score CFG: 3.0; fake-score CFG: 0.0 \\
Timestep range and shift
& 1,000 training timesteps; shift factor 5.0 \\
Update schedule
& Five fake-score updates per generator update;
discriminator updated alongside fake score \\
Optimizer
& AdamW; $\beta_1=0$, $\beta_2=0.999$; weight decay 0.01 \\
Learning rates
& Generator: $2\times10^{-6}$;
fake score and discriminator heads: $4\times10^{-7}$ \\
Batch size
& 1 per GPU on 16 GPUs; four gradient accumulation steps \\
EMA & Decay 0.99, starting at step 200 \\
GAN branch
& Layers $\{13,21,29\}$ (zero-based); 3 registers;
1,536 feature dim; 8,192 FFN dim \\
Discriminator outputs
& 11 logits: 4 for real-distribution alignment,
3 for inter-step consistency alignment,
and 4 for first-frame enhancement \\ 
GAN loss
& Relativistic logistic loss;
RA and CA weights: 1; FE weight: 0.05 \\
ODE supervision
& Weight 0.05; early 200-step warm-up \\
Discriminator regularization
& R1 and R2 weights: 20; perturbation standard deviations: 0.05 \\
Systems
& Mixed precision, gradient checkpointing, hybrid full-shard FSDP \\
Training steps & 1,000 iterations \\
\bottomrule
\end{tabular}
\end{table*}

\noindent\textbf{Long-Video Inference and Evaluation.}
Long-video generation uses 123 / 243 latent frames, decoded into
489 / 969 video frames and exported at 16 FPS for 30s / 60s.
We retain three global sink frames, two long-term memory
frames, and one recent frame during generation.
Memory is maintained during the initial 21 latent frames
and used for attention thereafter.
Memory selection uses clean historical queries and keys
before RoPE, excluding the current noisy query.
Section~\ref{sec:noisy_query} evaluates this choice.
Position Mapping permits multiple memory frames to
share a temporal position, as analyzed in
Section~\ref{sec:position_collision}.
Main evaluations use 946 prompts with five samples per prompt,
while long-video ablations use a fixed 128-prompt subset.
Throughput is measured on a single GPU after an unmeasured
warm-up, excluding VAE decoding and video encoding.

\noindent\textbf{Overall Process of Any Forcing.}
To facilitate intuitive understanding, we provide the pseudocode as shown in Algorithm~\ref{alg1} (training) and Algorithm~\ref{alg2} (inference).

\newtcolorbox[auto counter]{AnyForcingBox}[2]{%
  colback=black!5,
  colframe=black!30,
  colbacktitle=black!20,
  coltitle=black,
  fonttitle=\small\bfseries,
  fontupper=\footnotesize,
  title={Algorithm~\thetcbcounter: #1},
  label={#2},
  boxrule=0.7pt,
  arc=1.5mm,
  left=2mm,right=2mm,top=2mm,bottom=2mm,
  toptitle=1.5mm,bottomtitle=1.5mm,
  equal height group=anyforcing-pseudocode-pair,
  valign=top
}

\begin{figure*}[h]
\centering
\begin{minipage}[t]{0.485\textwidth}
\vspace{0pt}
\begin{AnyForcingBox}{Training Process of Any Forcing}{alg1}
\begin{algorithmic}[1]
\Require Student $G_\theta$, frozen teacher $\Phi$,
fake-score model $F_\phi$, discriminator $D_\psi^{\mathrm{cls}}$;
training iterations $S_{\mathrm{train}}$, maximum steps $K$,
update ratio $r$, warm-up length $W$
\For{$n=0,\ldots,S_{\mathrm{train}}-1$}
    \State Sample real latents $\boldsymbol{x}_0^{\mathrm{real}}$,
    text $\boldsymbol{c}$, and sampling steps $k$
    \State Sample initial noise $\boldsymbol{z}$
    \If{$n<W$ and $k=1$}
        \State Load paired teacher ODE endpoints
        $(\boldsymbol{x}_T^{\mathrm{tgt}},\boldsymbol{x}_0^{\mathrm{tgt}})$
        \State $\boldsymbol{z}\gets\boldsymbol{x}_T^{\mathrm{tgt}}$
    \EndIf
    \State Generate $\widehat{\boldsymbol{x}}_0^{(k)}$
    using causal rollout
    \If{$k<K$}
        \State Generate $\widehat{\boldsymbol{x}}_0^{(k+1)}$
        from $z$ without gradient
    \Else
        \State $\mathcal{L}_G^{\mathrm{CA}}\gets0,\quad
        \mathcal{L}_D^{\mathrm{CA}}\gets0$
    \EndIf
    \State Select task-specific logits using task
    and sampling-step labels
    \If{$n\bmod(r+1)<r$}
        \State Update $\phi$ with $\mathcal{L}_{\mathrm{fake}}$
        on generated samples $\widehat{\boldsymbol{x}}_0^{(k)}$
        \State Update $\psi$ with
        $\mathcal{L}_D^{\mathrm{RA}}+\mathcal{L}_D^{\mathrm{CA}}
        +\lambda_{\mathrm{FE}}\mathcal{L}_D^{\mathrm{FE}}$
    \Else
        \State Compute $\mathcal{L}_{\mathrm{DMD}}$
        using $\Phi$ and $F_\phi$
        \State $\mathcal{L}_{\mathrm{adv}}\gets
        \mathcal{L}_G^{\mathrm{RA}}+\mathcal{L}_G^{\mathrm{CA}}
        +\lambda_{\mathrm{FE}}\mathcal{L}_G^{\mathrm{FE}}$
        \State $\mathcal{L}_{\mathrm{ODE}}\gets0$
        \If{$n<W$ and $k=1$}
            \State $\mathcal{L}_{\mathrm{ODE}}\gets
            \|\widehat{\boldsymbol{x}}_0^{(1)}
            -\boldsymbol{x}_0^{\mathrm{tgt}}\|_2^2$
        \EndIf
        \State Update $\theta$ with
        $\mathcal{L}_{\mathrm{DMD}}+\mathcal{L}_{\mathrm{adv}}
        +\lambda_o\mathcal{L}_{\mathrm{ODE}}$
        \State $\theta_{\mathrm{ema}}\gets\mu\theta_{\mathrm{ema}}+(1-\mu)\theta$, when EMA is activated
    \EndIf
    \EndFor
\Output Optimized Generator with parameters $\theta_{\mathrm{ema}}$
\end{algorithmic}
\end{AnyForcingBox}
\end{minipage}\hfill
\begin{minipage}[t]{0.485\textwidth}
\vspace{0pt}
\begin{AnyForcingBox}{Inference Process of Any Forcing}{alg2}
\begin{algorithmic}[1]
\Require Generator $G_\theta$, text $\boldsymbol{c}$,
latent-frame count $F$;
sink size $S$, memory capacity $M$,
recent queue capacity $N$,
trained temporal horizon $L$
\State Initialize KV cache; initialize $\mathcal{S},\mathcal{M},\mathcal{R}$ only for $F>L$
\State Sample initial noise sequence $\boldsymbol{z}\sim\mathcal{N}(\boldsymbol{0},\boldsymbol{I})$
\For{$i=0,\ldots,F-1$}
    \If{$F>L$ and $\mathcal{R}$ contains $N$ frames}
        \State $C\gets\operatorname{PopOldest}(\mathcal{R})$
        \State $\mathcal{M}\gets$ \Call{UpdatePool}{$C,\mathcal{M},\mathcal{R}$}
    \EndIf
    \State Use the original attention context for $i<L$; otherwise, attend to $\mathcal{S}$, $\mathcal{M}$, and $\mathcal{R}$ using the position mapping in Eq.~\eqref{eq:memory_remapping}.
    \State Set current-frame position to $i$ for $i<L$, otherwise $L-1$
    \State Generate $\widehat{\boldsymbol{x}}_0^i$ from frame noise $\boldsymbol{z}_i$ with $1^*$ or $1^\dagger$ configuration
    \State Clean-forward frame $i$ and update its KV cache
    \State If $F>L$, cache clean states in
    $\mathcal{S}$ ($i<S$) or $\mathcal{R}$ ($i\geq S$)
\EndFor
\State \textbf{return} $\widehat{\boldsymbol{x}}_0$
\Output Generated video latents $\widehat{\boldsymbol{x}}_0$
\Statex
\Function{UpdatePool}{$C,\mathcal{M},\mathcal{R}$}
    \State \textbf{if} the pool is not full,
    \textbf{return} $\mathcal{M}\cup\{C\}$
    \State Retrieve clean historical Q/K summaries before RoPE
    \State Compute sink semantic relevance $s_j$
    using Eq.~\eqref{eq:global_relevance}
    \State $A\gets\operatorname*{arg\,min}_{j\in\mathcal{M}}s_j$
    \State \textbf{if} $s_C\leq s_A$,
    \textbf{return} $\mathcal{M}$
    \State Compute $a(j,A),a(j,C)$ for $j\in\mathcal{R}$
    using Eq.~\eqref{eq:summary_response}
    \State $(p_A^{(j)},p_C^{(j)})\gets
    \operatorname{softmax}([a(j,A),a(j,C)])$
    \State $U_v\gets\frac{1}{N-1}
    \sum_{j\in\mathcal{R}}p_v^{(j)}$, $v\in\{A,C\}$
    \State \textbf{if} $U_C>U_A$,
    $\mathcal{M}\gets(\mathcal{M}\setminus\{A\})\cup\{C\}$
    \State \textbf{return} $\mathcal{M}$
\EndFunction
\end{algorithmic}
\end{AnyForcingBox}
\end{minipage}
\end{figure*}

\section{Human Preference Details}
\label{sec:human_preference_details}

\noindent\textbf{Study protocol.}
We conduct a blind pairwise user study with 48 evaluators to
complement the automated evaluation. For short-video generation,
we compare Any Forcing with ASD and One Forcing; for long-video
generation, we compare it with Deep Forcing and Infinity RoPE.
Each comparison presents two videos generated from the same text
prompt under the same sampling configuration. The videos are
displayed as A and B, with their left--right order randomized.
Neither the source methods nor the first-frame sampling budgets
are disclosed to participants.

\noindent\textbf{Sampling and presentation.}
Each questionnaire contains 15 comparisons sampled without
replacement from a preselected case pool. A case is shown at most
once within a questionnaire, and one available comparison between
Any Forcing and a baseline is randomly selected for that case.
Short- and long-video cases, as well as the two first-frame
sampling configurations, are mixed without fixed per-participant
quotas.

\noindent\textbf{Evaluation instructions and interface.}
Participants are instructed to select the video they prefer
overall, considering both visual quality and semantic alignment
with the prompt. Visual quality includes clarity, structural
plausibility, motion smoothness, and temporal consistency;
semantic alignment concerns whether the depicted content and
actions match the prompt. Participants must choose either A or B;
no tie option is provided. The instruction is:

\begin{quote}
Watch both videos and select the one that is better overall,
considering visual quality and consistency with the prompt.
Base your judgment on the displayed videos and the prompt.
If the difference is small, select the video you slightly prefer.
\end{quote}

The browser-based interface supports synchronized playback,
pausing, and seeking, allowing participants to replay and inspect
the same temporal segment in both videos. Confirmation is enabled
only after both videos have started playing. No method-specific
quality assessments or selection rationales are shown.

\begin{figure*}[h]
    \centering
    \includegraphics[width=\textwidth]{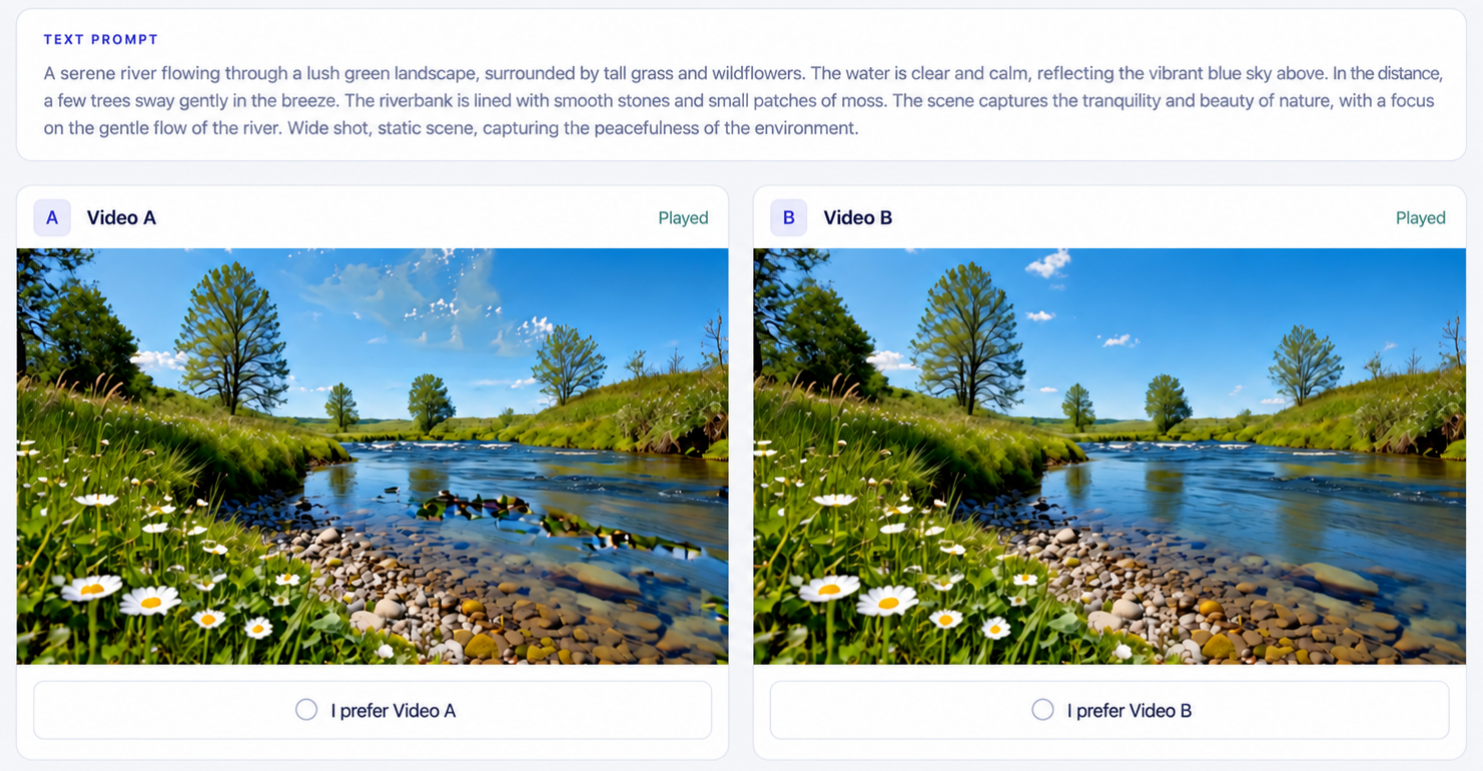}
    \caption{Interface for the blind pairwise user study.
    Participants compare videos generated from the same prompt
    and select their preferred result based on visual quality
    and semantic alignment.}
    \label{fig:user_study_interface}
\end{figure*}

\noindent\textbf{Preference aggregation.}
We report the preference rate of Any Forcing separately for each
baseline and sampling configuration. The $1^{*}$ and
$1^{\dagger}$ configurations use four and two denoising steps for
the first latent frame, respectively, followed by one step for
subsequent frames. For baseline $b$ and configuration $s$, the
preference rate is computed as
\begin{equation}
    \mathrm{WinRate}_{b,s}
    =
    \frac{N^{\mathrm{Any\ Forcing}}_{b,s}}
         {N^{\mathrm{valid}}_{b,s}},
\end{equation}
where $N^{\mathrm{Any\ Forcing}}_{b,s}$ is the number of valid
comparisons in which Any Forcing is preferred, and
$N^{\mathrm{valid}}_{b,s}$ is the total number of valid comparisons
for that baseline and configuration. Only complete questionnaires
are included, and repeated exports of the same questionnaire are
counted once. The visual interface is presented in Fig.~\ref{fig:user_study_interface}.

\noindent\textbf{Participation and evaluation scope.}
Participation is voluntary, and participants may pause or stop
at any time. The questionnaire does not request names or email
addresses; responses are associated with anonymous session
identifiers. The case pool combines manually curated examples
and randomly drawn examples. Accordingly, the reported preference
rates characterize the evaluated case pool rather than an
unbiased estimate over all generated videos.

\section{Effect of ODE Endpoint Warm-Up.}\label{ode_warm}
\begin{figure*}[t]
    \centering
    \includegraphics[width=\textwidth]{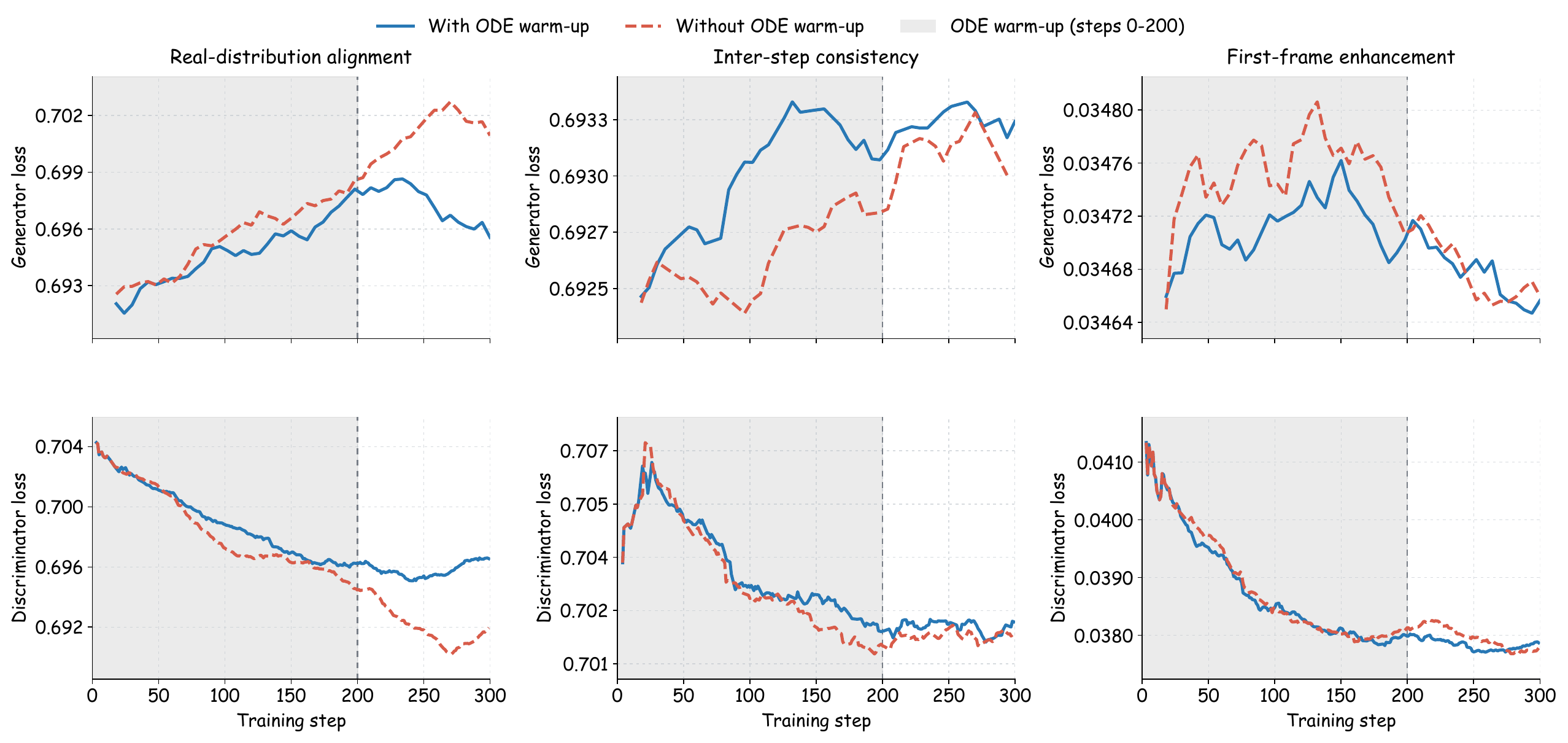}
    \caption{Adversarial training dynamics with and without ODE
    endpoint warm-up over the first 300 training steps.
    Columns show real-distribution alignment,
    inter-step consistency, and first-frame enhancement;
    rows show generator and discriminator losses, respectively.
    The shaded region indicates the
    200-step warm-up stage.}
    \label{fig:ode_gan_stability}
\end{figure*}

Fig.~\ref{fig:ode_gan_stability} compares the generator and discriminator losses across the three adversarial tasks over the first 300 training steps.
The clearest difference appears in real-distribution alignment after ODE supervision ends at step 200: without ODE warm-up, the generator loss increases substantially and the discriminator loss declines more sharply, whereas both change more gradually with warm-up.
It should be noted that these loss fluctuations reflect early training stability but do not directly indicate the magnitude of adversarial gradients. The ablation results in Table~\ref{tab:ablation} further demonstrate the effectiveness of early ODE endpoint supervision.

\section{Additional Analysis of Positional Mapping and Memory Selection}
\label{sec:memory_analysis}

\subsection{Current Noisy Queries}
\label{sec:noisy_query}

Our local value contribution criterion uses clean historical
queries and excludes the current noisy query.
To examine whether the latter provides complementary
information, we add the current frame's query from its
first denoising forward pass to the local reference queries.
All representations are taken before RoPE.
Each query independently compares the same candidate and
memory entry under a joint softmax, after which their
attention weights are averaged.

\begin{figure*}[t]
    \centering
    \includegraphics[width=\textwidth]{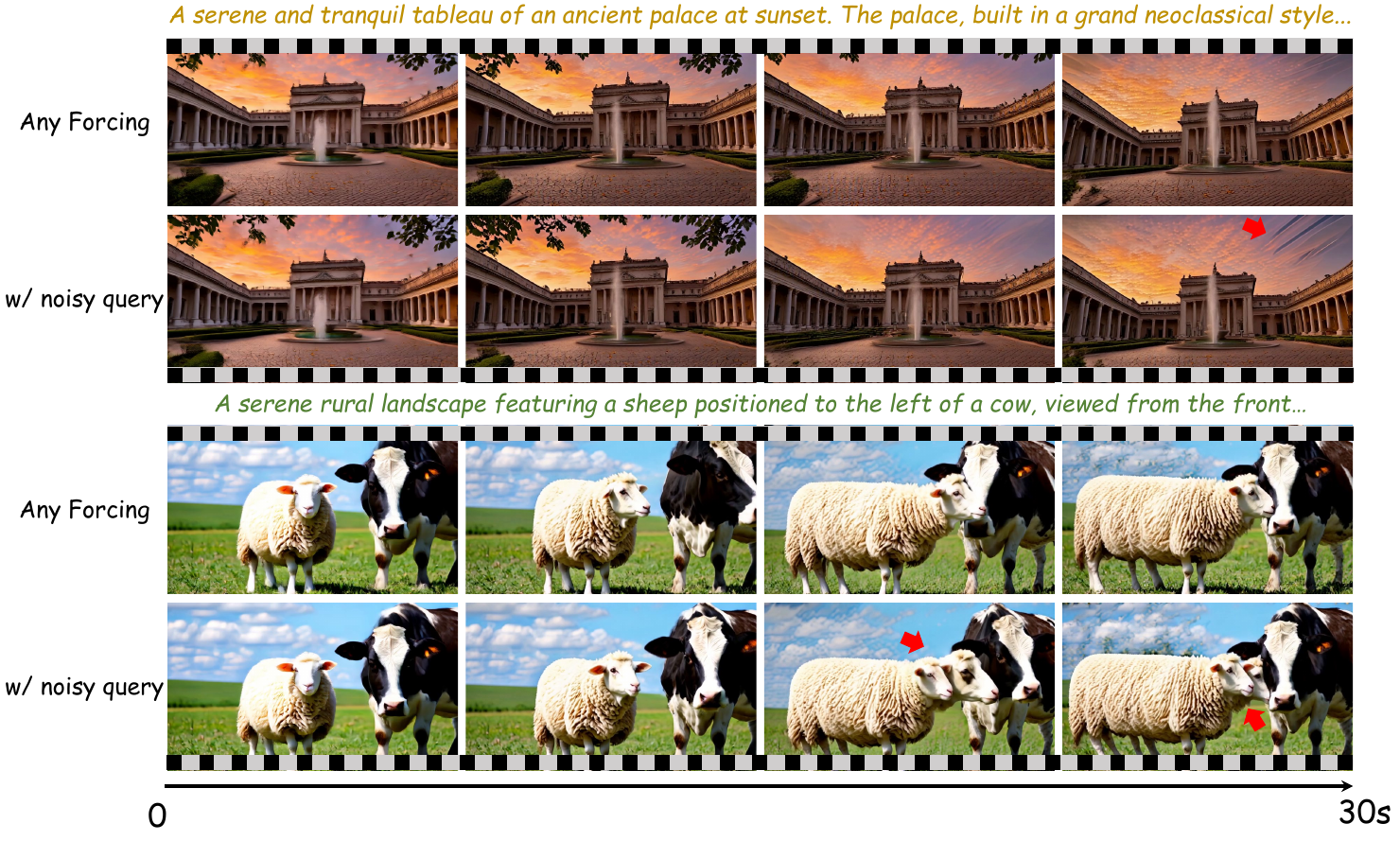}
    \caption{Qualitative examples of generation with the current noisy query included in local value evaluation.
    Red arrows indicate regions of content degradation, including duplicated sheep heads.}
    \label{fig:noisy_query}
\end{figure*}

Table~\ref{tab:noisy_query} shows mixed results.
Including the noisy query improves spatial relationship
and scene scores but reduces object recognition,
multiple-object accuracy, and temporal style.
Imaging quality and temporal flickering change only slightly.
The qualitative examples in Fig.~\ref{fig:noisy_query}
show structural artifacts when the current noisy query
is included, such as duplicated sheep heads.
These examples highlight potential degradation of
subject integrity.

\begin{table}[h]
\centering
\caption{Effect of including the current noisy query
in local value evaluation.}
\label{tab:noisy_query}
\setlength{\tabcolsep}{1.5pt}
\renewcommand{\arraystretch}{1.1}
\small
\begin{tabular}{l*{8}{c}}
\toprule
Setting & IQ & AQ & OC & MO & SR & TF & Scene & TS \\
\midrule
\rowcolor{gray!20}
\textbf{Any Forcing}
& 71.71 & \textbf{59.38}
& \textbf{93.50} & \textbf{82.08}
& 89.14 & \textbf{99.34}
& 59.70 & \textbf{23.28} \\
w/ noisy query
& \textbf{71.74} & 59.36 & 93.31 & 81.84
& \textbf{89.77} & \textbf{99.34}
& \textbf{59.88} & 23.10 \\
\bottomrule
\end{tabular}
\end{table}

\subsection{Positional Collisions}
\label{sec:position_collision}

Positional rescaling compresses an expanding historical
timeline into a fixed temporal encoding horizon.
As generation proceeds, multiple original frame positions
may therefore map to the same discrete position.
Such collisions preserve nondecreasing chronological order,
but reduce temporal separability: the affected frames share
a temporal RoPE phase while retaining distinct
content-dependent keys and values.
The capacity constraint ensures sufficient positions for
the retained frames, but does not guarantee collision-free
proportional rescaling.

From the perspective of temporal span, adjacent historical
frames may be close relative to the entire compressed
timeline.
Assigning them the same position represents this small
separation at a coarser temporal resolution.
Forcing distinct positions instead introduces a minimum
encoded separation, which can distort the relative
distances obtained by proportional rescaling.
We therefore permit collisions while preserving the
chronological order of retained memory.

To evaluate this choice, we enforce strictly increasing
memory positions in an ablation.
Colliding entries are shifted forward in chronological
order, with space reserved for subsequent entries to
respect the memory interval boundary.
For example, $(9,9)$ becomes $(9,10)$; near an upper
boundary of 18, three entries are assigned $(16,17,18)$.
All other settings remain identical.

Table~\ref{tab:position_collision} shows that preventing
collisions reduces performance on most evaluated metrics.
These results suggest that strict temporal separation is
not necessary for the retained memory and support the
default proportional mapping with permitted collisions.

\begin{table}[h]
\centering
\caption{Effect of preventing positional collisions.}
\label{tab:position_collision}
\setlength{\tabcolsep}{1.5pt}
\renewcommand{\arraystretch}{1.1}
\small
\begin{tabular}{l*{8}{c}}
\toprule
Setting & IQ & AQ & OC & MO & SR & TF & Scene & TS \\
\midrule
\rowcolor{gray!20}
\textbf{Any Forcing}
& \textbf{71.71} & \textbf{59.38}
& \textbf{93.50} & 82.08
& \textbf{89.14} & \textbf{99.34}
& \textbf{59.70} & \textbf{23.28} \\
w/o collisions
& 71.65 & 59.34 & 92.90 & \textbf{82.09}
& 89.03 & 99.31 & 58.95 & 23.13 \\
\bottomrule
\end{tabular}
\end{table}

\section{More Results}\label{more_res}
\subsection{VBench Scores Across All Dimensions}

\begin{figure*}[h]
    \centering
    \includegraphics[width=\textwidth]%
    {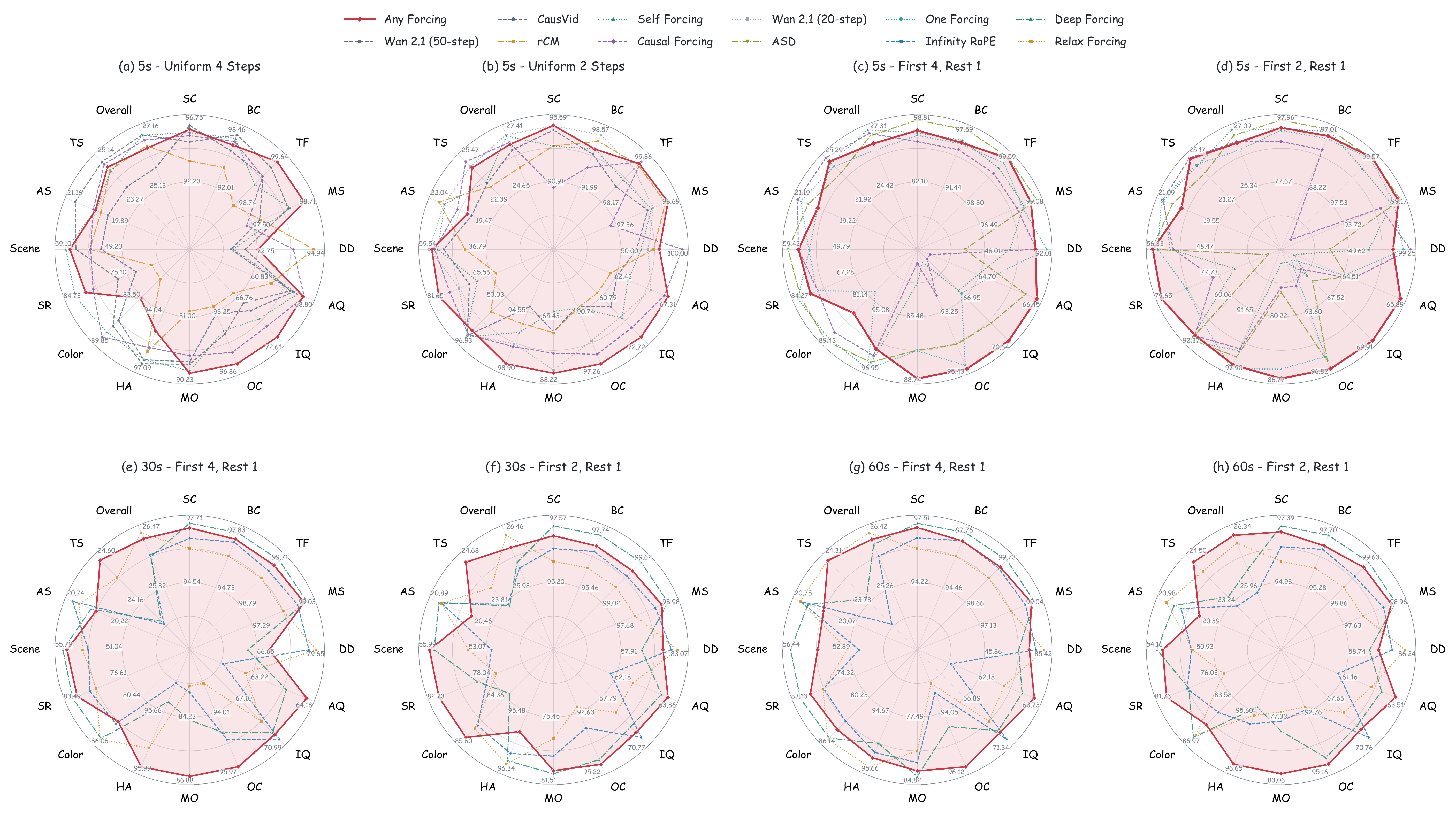}
    \caption{Comparison across all 16 VBench dimensions.
    Top: 5s generation with uniform four-step and two-step
    sampling, followed by one-step generation with four
    and two first-frame steps.
    Bottom: corresponding one-step comparisons for
    30s (left two panels) and 60s (right two panels).
    Any Forcing is highlighted in red. The radar plots use axis-specific display ranges shared by all methods within each panel; outermost labels denote axis limits rather than method scores.
    SC: Subject Consistency;
    BC: Background Consistency;
    TF: Temporal Flickering;
    MS: Motion Smoothness;
    DD: Dynamic Degree;
    AQ: Aesthetic Quality;
    IQ: Imaging Quality;
    OC: Object Class;
    MO: Multiple Objects;
    HA: Human Action;
    SR: Spatial Relationship;
    AS: Appearance Style;
    TS: Temporal Style;
    Overall: Overall Consistency.}
    \label{fig:vbench_radar}
\end{figure*}

Fig.~\ref{fig:vbench_radar} provides a detailed comparison
across video lengths and sampling schedules.
For 5s one-step generation, Any Forcing achieves the
highest aesthetic quality, imaging quality, object-class,
and multiple-object scores under both first-frame settings,
highlighting improvements in visual fidelity and semantic
correctness.
Its strong performance under uniform multi-step sampling
further supports its sampling-step generality.
For both 30s and 60s generation, Any Forcing consistently
leads in aesthetic quality, object class, and temporal style
under both first-frame settings.
With two first-frame steps, it also achieves the strongest
multiple-object and spatial-relationship scores at 60s.
These results suggest that adaptive memory selection
helps retain useful historical information for long-video
generation.
Overall, the comparison reveals complementary gains
in perceptual quality and semantic fidelity across
sampling settings and generation lengths.

\subsection{More Visualizations}

\begin{figure*}[t]
    \centering
    \includegraphics[width=\textwidth]%
    {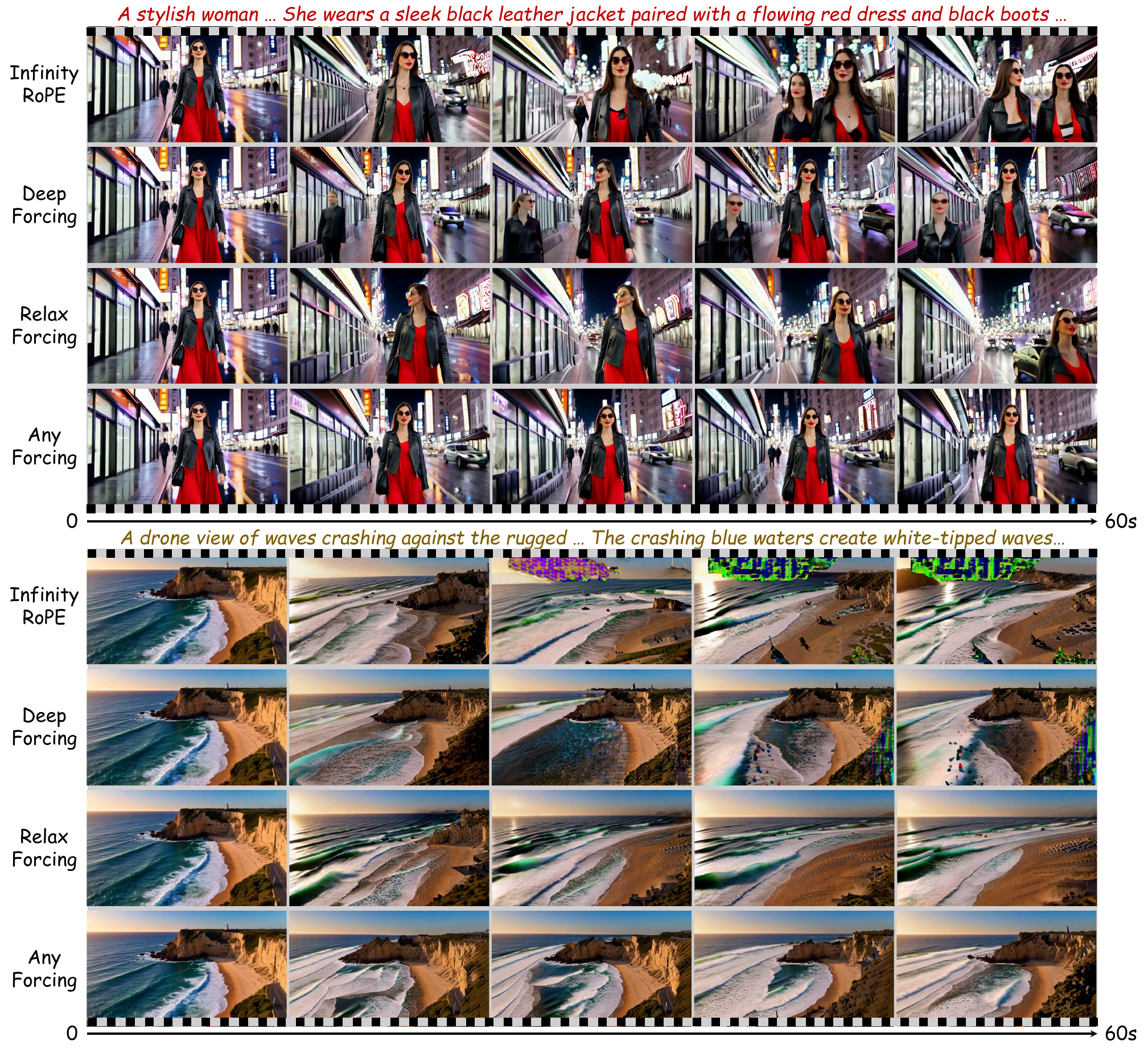}
    \caption{Comparisons of 60-second video
    generation on a Tokyo
    street scene (top) and a coastal landscape (bottom). All methods adopt the $1^{*}$ sampling configuration.}
    \label{fig:long_video_qualitative}
\end{figure*}

\noindent\textbf{Comparison of 60-second videos.} Fig.~\ref{fig:long_video_qualitative} compares
long-video generation in scenes with different
appearance and structural requirements.
In the Tokyo example, Any Forcing maintains the
subject's characteristic black jacket, red dress,
and sunglasses, together with relatively stable
subject framing.
Competing results exhibit varying degrees of
clothing variation and framing drift.
In the coastal example, Infinity RoPE develops
pronounced grid-like artifacts in the sky,
while Deep Forcing exhibits patterned distortions
in the water and surrounding terrain.
Relax Forcing gradually loses the prominent cliff
structure present in the initial frame.
In comparison, Any Forcing better preserves the
recognizable coastline and cliff geometry
throughout the sampled sequence.
These examples complement the quantitative results,
suggesting that adaptive long-term memory selection
helps preserve both scene structure and subject
appearance during one-step generation.

\begin{figure*}[t]
    \centering
    \includegraphics[width=\textwidth]{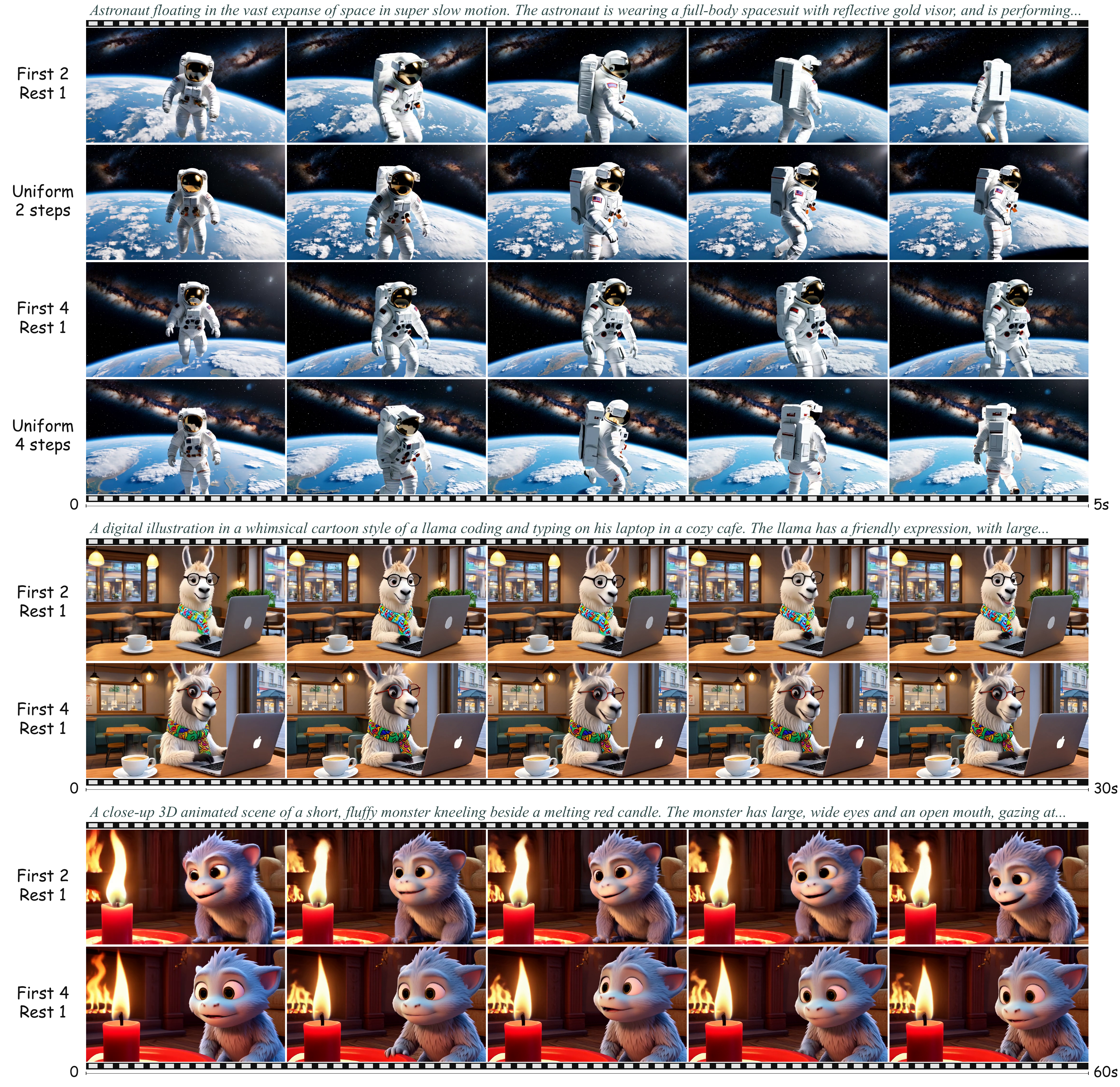}
    \caption{Comparisons of Any Forcing under different
    sampling schedules.
    The top, middle, and bottom groups show 5-, 30-, and
    60-second generation, respectively.}
    \label{fig:sampling_comparison}
\end{figure*}

\noindent\textbf{Comparison of different sampling strategies.}
As shown in Fig.~\ref{fig:sampling_comparison},
Any Forcing produces visually coherent results across
different sampling schedules and video lengths.
For 5-second generation, both uniform few-step sampling
and first-frame-enhanced one-step sampling preserve
recognizable astronaut structure and the space environment,
although poses and framing vary across schedules.
For 30-second generation, the character's glasses,
colorful scarf, and surrounding cafe remain largely
consistent throughout the sampled frames.
The 60-second examples similarly retain the character's
appearance, the red candle, and the overall scene layout
under both first-frame settings.
These examples qualitatively support the any-step
generality of Any Forcing and its ability to maintain long-video consistency with one-step sampling strategy.